\documentclass[manuscript]{acmart}

\AtBeginDocument{%
  \providecommand\BibTeX{{%
    \normalfont B\kern-0.5em{\scshape i\kern-0.25em b}\kern-0.8em\TeX}}}

\setcopyright{none}

\usepackage{comment}

\begin{document}

\title{Soft Robotic Technological Probe for Speculative Fashion Futures}


\author{Amy Ingold}
\orcid{1234-5678-9012}
\email{amy.ingold@bristol.ac.uk}
\author{Loong Yi Lee}
\orcid{0000-0002-7354-6018}
\author{Richard Suphapol Diteesawat}
\orcid{0000-0001-7002-5658}
\author{Ajmal Roshan}
\orcid{0000-0002-0655-1941}
\author{Yael Zekaria}
\orcid{0000-0002-2806-1206}
\author{Edith-Clare Hall}
\orcid{0000-0003-4492-3416}
\author{Enrico Werner}
\orcid{0009-0006-3682-3078}
\author{Nahian Rahman}
\orcid{0000-0002-5165-2821}
\author{Elaine Czech}
\orcid{0000-0002-8923-0085}
\author{Jonathan M Rossiter}
\orcid{0000-0002-9109-9987}
\email{jonathan.rossiter@bristol.ac.uk}
\affiliation{%
  \institution{University of Bristol}
  \streetaddress{Beacon House, Queens Road}
  \city{Bristol}
  \country{UK}
  \postcode{BS8 1QU}
}




\renewcommand{\shortauthors}{Ingold, et al.}

\begin{abstract}
Emerging wearable robotics demand design approaches that address not only function, but also social meaning. In response, we present Sumbrella, a soft robotic garment developed as a speculative fashion probe. We first detail the design and fabrication of the Sumbrella, including sequenced origami-inspired bistable units, fabric pneumatic actuation chambers, cable driven shape morphing mechanisms, computer vision components, and an integrated wearable system comprising a hat and bolero jacket housing power and control electronics. Through a focus group with twelve creative technologists, we then used Sumbrella as a technological probe to explore how people interpreted, interacted, and imagined future relationships with soft robotic wearables. While Sumbrella allowed our participants to engage in rich discussion around speculative futures and expressive potential, it also surfaced concerns about exploitation, surveillance, and the personal risks and societal ethics of embedding biosensing technologies in public life. We contribute to the Human-Robot Interaction (HRI) field key considerations and recommendations for designing soft robotic garments, including the potential for kinesic communication, the impact of such technologies on social dynamics, and the importance of ethical guidelines. Finally, we provide a reflection on our application of speculative design; proposing that it allows HRI researchers to not only consider functionality, but also how wearable robots influence definitions of what is considered acceptable or desirable in public settings.


\begin{figure}[h]
\includegraphics[width=9cm]{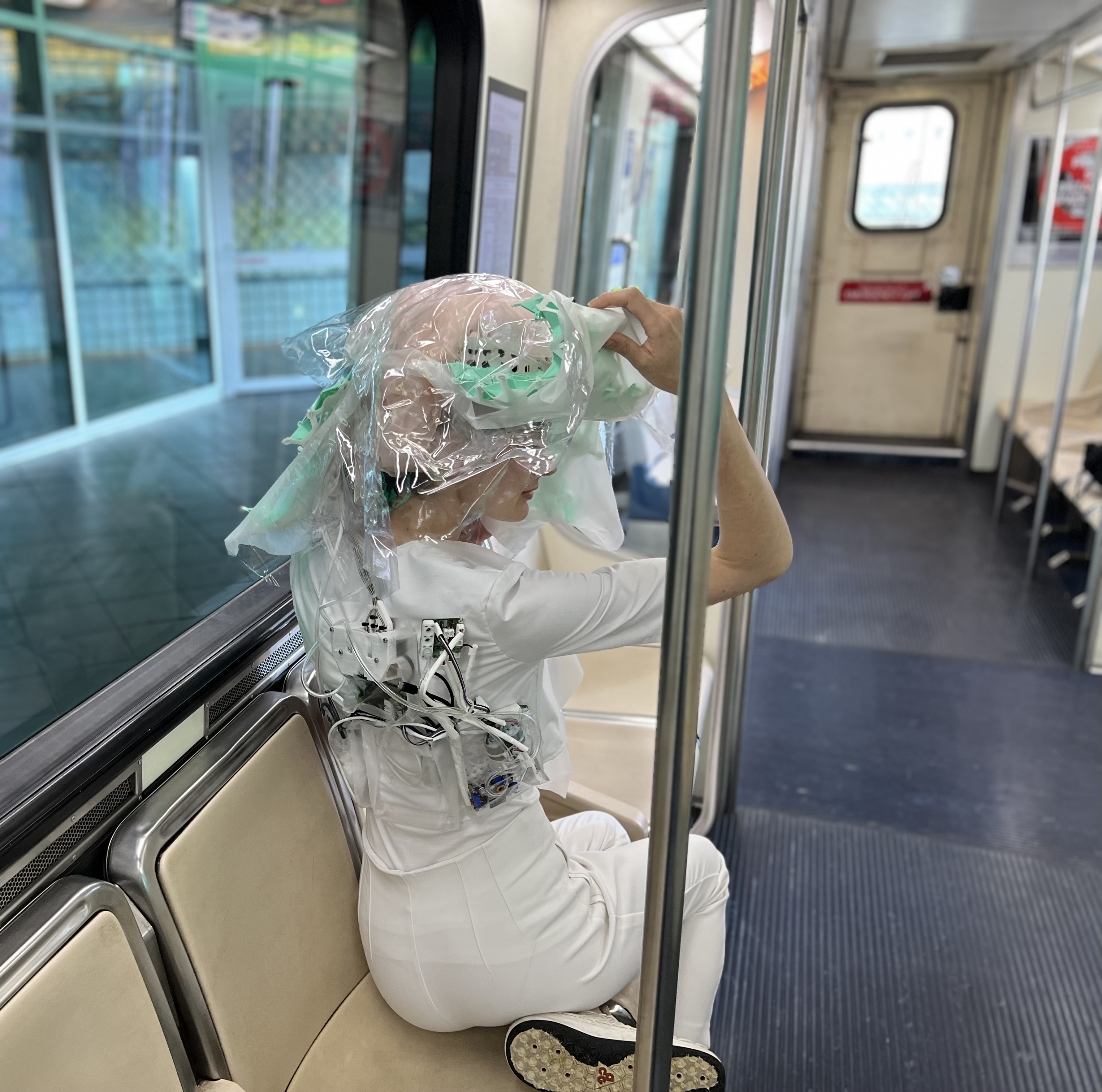}
\caption{The Sumbrella robotic hat, being worn while riding a train.}
\Description{A person sitting on a train wearing a robot hat and a jacket with translucent pockets through which cables and other inner workings are visible.}
\label{fig_sumbrella_worn}
\end{figure}

\end{abstract}






\maketitle

\section{Introduction}

Connection with others is essential for our well-being, deeply rooted in our need for social interaction and belonging \cite{klussman2020connection}. However, in our current society, building and maintaining social connections can be difficult without intervention (i.e., needing social applications to form romantic relationships \cite{lefebvre2018swiping}). One simple way we can create an opportunity for connection is through what we wear. We use our clothes and accessories to signal to others, i.e., to stand out from, or blend into, a crowd. Using our clothing in this way is well understood in the realm of fashion. Employing semiotics, the study of signs and sign-using behaviour, to fashion, reveals the way in which we signify specific social and cultural information about ourselves \cite {karunaratne2018exploring}. While this form of communication allows broad statements about our affiliations, personalities and intentions, clothing is a relatively static temporal signal. This stasis limits the representation of the nuances of our changing internal states and preferences. 

However, integration of technology into clothing and accessories allows more dynamic and representational fashion design \cite{schaar2021predicting}. Research into clothing as an interface between wearer and onlooker, both as a means to facilitate social interaction and to preserve personal space, has only in the recent decade received attention by wearable technology designers \cite{hrga2019wearable, Jabari16112024}. In these spaces, emerging wearable technologies border the science fiction of fashion by allowing the pieces to react on the body to specific stimuli (i.e., changing shape \cite{Gonzalez_2016} or emitting smoke \cite{SmokeDress2012} in response to the viewer's gaze \cite{Howarth_2014}, displaying the wearer’s biosignals \cite{jeong_2013, saini2025inflatedata} or the proximity of a bystander \cite{hrga2019wearable}).

Soft robots, with their natural and fluid movements, represent a significant opportunity for behavioural communication. Unlike rigid machines, these soft robotics systems can mimic the flexibility and adaptability of biological organisms, making them ideal for a wide range of industrial applications \cite{rossiter2016soft, elango2015review}. However, the potential of soft robotics extends far beyond industrial applications, offering new possibilities to expand fields like wearable technology. The ability of soft robotics to move in a more natural, organic-like manner enables softer and safer physical interactions, reducing the risk of injury during use \cite{laschi2016soft, pfeifer2012challenges}. This tactile quality is critical in applications where robots must operate in close proximity to humans, providing a more comfortable and intuitive interface \cite{arnold2017tactile}.

Investigations into integrating soft robotics into clothing have already begun, with application including physical assist and space travel. These offer muscle-assist capabilities; enhancing, or replacing lost mobility and strength for the wearer, or maintaining strength in low-gravity environments \cite{zhu2022soft, yumbla2021human, porter2020soft, o2017soft, pulvirenti2022towards}. Although the concept of soft robots as wearable garments is still in its nascent stages and not yet prevalent in mainstream fashion, their inherent tactility, safety, and adaptability make them exceptionally well-suited for this purpose. Additionally, given their fluid and natural appearance, soft robotic wearables have the potential to augment communication between the wearer and others, unlike computational wearable technology. This allows soft robotic wearables to help the wearer convey their intentions through movement and form, similar to how a dog might for its owner \cite{somppi2022dog}. This study explores the potential of a soft robotic wearable in the context of communication and protection, examining how this technology can enhance human interaction through biomimetic kinesic communication\footnote{Kinesics is the interpretation of body communication and nonverbal behaviour related to movement of any part of the body or the body as a whole.} mechanisms.

This paper engages the kinesic potential of soft robotics via a technology probe to speculate on the future of fashion from twelve creative technologists. We aimed to probe the future of soft robotics in fashion by answering the following research questions:

\begin{description}
    \item \textbf{RQ1:} To what extent do people trust the automation of social intent by computational reading of biosignals?
    \item \textbf{RQ2:} Do people understand social meaning from novel soft robotic kinesics?
    \item \textbf{RQ3:} What are the social and ethical implications of delivering complex robotic wearables for widespread use?
\end{description}

In pursuit of this goal, we introduce Sumbrella (Figure \ref{fig_sumbrella_worn}), a soft robotic hat designed to explore aesthetic and speculative futures. Sumbrella is meant to not only exemplify an innovative application of soft robotics within the fashion space but also serve as a technological probe to provoke discussion on speculative futures. The findings from our thematic analysis not only provide an evaluation of our design but also reveal essential ethical implications of soft robotic fashion futures. This exploration offers insights into how speculative design helps expand the bounds of future spaces for robots.


\section{Related Work}

Our work focuses on designing soft robotics as a new fashion category. Here we provide an overview of how fashion and computational wearables have allowed and even helped wearers express themselves. We then discuss how robots, particularly soft robots, have their own aesthetic and kinesic abilities that, if incorporated into wearables, could add a different dynamic, less explored in computational wearables. We conclude this section by discussing how speculative design could help us to reimagine how we engage with soft robotics in everyday life. 

\subsection{Wearable Expression}
Fashion as a form of expression can be used to invite conversation or debate (i.e., Lady Gaga's meat dress \cite{alexander2011}) or block out the world (i.e., Ostrich's Original Napping Pillow \cite{ostrich}). In the world of fashion, physical barriers are one means to non-verbally convey a wearer's needs or wants. A stiff hoop skirt creates a physical barrier to tell potential suitors it is ok to look but not touch \cite{wilson2021women}. Other similar strategies include: adapting designs to include positioning around the waist for gender-neutral wearing and adding to swimwear for defence in all situations \cite{Multiply_2020}. This type of signalling has also been achieved mechanically \cite{Quah_2015, Intravaia_2020}, and by employing pneumatics. For instance, designer Liu \cite{Art_2021} felt that dressing appropriately created feelings of disappearing and designed inflatable pieces that purposely took up space in the physical and the social realm to create non-conformity. In contrast, designer Gao used fingerprint technology to invite the wearer to connect with strangers \cite{Levy_2017}. Gao's piece encouraged strangers to come up close to the wearer as the dress would become animated only for those who had not interacted with it yet. This design more obviously took away some agency from the wearer as it was essentially others who controlled how their clothing would act. The critical point is that while fashion, like art, can allow the wearer to send a message, this message is not always straightforward and can be interpreted or manipulated by others. For instance, thinking back to the hoop skirt, as Wilson et al. \cite{wilson2021women} noted, while women saw them as sending a message of protection from sexual advances, men saw them as women inviting sex.

In contrast, the exploration of ubiquitous computing has meant that the design of wearable technology has focussed on creating tools which help with processing or communicating information quickly \cite{ometov2021survey}. Moving beyond simply smartwatches, computational wearables have in more recent years explored how wearable devices can express the wearer's information more playfully. This research specifically explores expressing a wearer's more hidden states (i.e., their emotions) and thus sees wearables as bodily extensions. One aim of bodily extending wearables is to help the wearer become more aware of their body. For instance, research by Saini et al. \cite{saini2025inflatedata} explored how a pneumatically actuated device worn on a bicep or calf could help bring better bodily awareness to the wearer by having the device inflate in correspondence to the amount of calories they have burned through exertion. Another aim of computational wearables as bodily extensions is to assist with communication and social interactions \cite{Buruk2023bodyexten, Wang2021Pneuskin}. For instance, researchers have utilised a pneumatic device on a wearer's body as a communication device. By employing an inflatable pouch to move body parts, it creates new social communications, i.e., by wiggling the ears to indicate the wearer is listening intently \cite{Saini2024pneu}. Or having the pneumatic device act as an amplifier, rendering subtle physiological changes perceivable by others. These inflatable devices have amplified the body's movements to facilitate social connection, either being worn or held to provide a haptic detection of another's breathing patterns \cite{Feng2024exbreath, Haynes2024just}. While this work begins to explore scaffolding social communication, the distinct opportunity for social interaction is underexplored in the area of pneumatically actuated wearables.

Some devices are more ambitious in their aim to generate opportunities for social interaction. 
The Wearable Aura, for example, combines proximity and inertial measurement unit (IMU) sensors with an EEG headset to render animations dependent on the wearer's cognitive status, environment and movement. These animations are projected around the wearer's personal space, encouraging others to interact with the wearer through the animations \cite{zheng2017wearable}. Devices such as this seek to augment the interim space between social actors. As Buruk et al. \cite{Buruk2023bodyexten} found, the more autonomous a worn entity seemed to be, or could be, can allow the device itself to become an object of social interaction. 

Still, as Jabari et al. \cite{Jabari16112024} noted, while computational wearables are moving toward incorporating more fashion aesthetics in their design, more multidisciplinary work with fashion experts is needed to help further these efforts. Other studies have approached this through workshops with fashion and engineering students \cite{genc2018exploring}, in contrast, our approach involved embedding these experts within the development team and evaluating the resulting prototype with public participants. By leveraging on the first author's eight years of experience in the fashion industry this research aims to explore how a medium like soft robotic design can better mirror fashion language when it comes to developing wearables for expression.

\subsection{Aesthetics and Kinesic Communication in Soft Robotics}

Another point Jabari et al. \cite{Jabari16112024} highlighted in their review of computational wearables is that kinetic transformations have the potential to change the future of fashion. However, despite the ability for kinetic movement to be dynamic within computational wearables, this ends up being limited to following a defined interaction sequence. By designing from the angle of soft robotics, this research aims to explore wearable technologies that are less scripted and controlled but are reactive and companion-like in their relationship to their human wearer, aiming instead to mimic kinesics, not just kinetics. 

Soft robotics offers a unique potential for enhancing social communication and interaction, especially through the use of soft, flexible materials that can autonomously mimic natural movements. Prior research into textile displays has highlighted how slowness, ambiguity, and material texture can shift the communicative role of wearable technologies away from functional screens and toward expressive, personal surfaces that better align with style and identity \cite{Devendorf2016i}. Similarly, soft robots can convey complex social signals more effectively than their rigid counterparts, thanks to their organic, fluid motions and tactile qualities \cite{sun2022softsar}. This has led to the development of the concept of ``\textit{soft biomorphism}'' \cite{christiansen2023brings}, where soft robots are designed to resemble soft-bodied organisms not directly, but through abstracted aesthetics that evoke feelings of well-being and comfort \cite{Christiansen2024ExSilico}. However, while these robots are often perceived as inviting and tactile, they can also elicit feelings of alienation or unpredictability due to their organic and sometimes chaotic appearances. This duality suggests that much remains to be explored in understanding how the aesthetics of soft robotics influence human perception and interaction \cite{christiansen2023brings}. 

Research indicates that soft robots elicit distinct interaction profiles, with people often wanting to touch or engage with them physically. For instance, tentacle-like appendages in soft robots have been found to provoke curiosity and encourage tactile exploration, further demonstrating the potential of these robots to prompt interaction \cite{jorgensen2022soft}. The integration of soft robots into wearable technology, such as the work on Nautica \cite{Belling_2021}, highlights the hybrid somatic relationships that can emerge between wearers and robots, driven by the shared materiality and behaviour of soft robotics.

Thus, soft robotics holds promise in the field of kinesics\footnote{The study of body language and gestures.} as a medium for conveying specific emotional states. For example, robots designed to replicate the movement of sea creatures or human organs, such as the heart and lungs, have successfully communicated emotions like joy or greetings, through their gestures \cite{farhadi2022exploring}. Additionally, research into simplistic, cylindrical soft robots has shown that expressive movements, such as bending, twisting, and expanding, can convey a wide range of emotions and social cues. The Sprout robot, for instance, was designed to generate a gentle and positive impression, using movements like bowing for greetings, leaning in for inquiries, and flopping forward to express sadness \cite{koike2024sprout, koike2024sprout2}.

This body of work demonstrates that soft robotics can play a significant role in enhancing social interactions through biomimetic designs and expressive gestures. The ability of these robots to evoke emotional responses and convey meaning through movement underscores their potential as tools for communication and interaction in various contexts, including wearable technologies that help negotiate social connection.

\subsection{Speculative Design in Fashion, Technology and Robotics}

Speculative design, when employed as a research method, allows the exploration of broad imaginative futures by moving away from the problem and solution-focused research strategies traditionally used, towards a focus on inquiry \cite{Galloway_2018, GamitoCaixas2023, Ringfort-Felner2025quality}. According to Dunne and Raby \cite{dunne2024speculative}, the application of speculative design is not meant to be an experiment that looks at how things are now but is meant to invite people to explore other possibilities altogether. It is through design that people can discuss, debate, and collectively define a more preferable future. Speculative design as design fiction has been used to critically examine the sociotechnical world that we live in \cite{taylor2004modern}. Harrington et al. \cite{harrington}, for instance, used elements of design fiction to encourage Black youths, who are often excluded from technology narratives, to speculate how they imagine technology would affect their future. Additionally, speculative design fiction has been employed to consider broader social landscapes such as technology's effects on urban environments \cite{forlano_2014, stals_2019} and more specific cases such as transhumanist design ideas\footnote{Designs which integrate machines with humans.} \cite{gen_2023} and exploring what to do when AI goes ethically wrong \cite{hanschke2024}. 

When it comes to exploring speculative design through the use of props, this is rarer since, as Dunne and Raby stated, a prop or object can only be classed as speculative if it was developed with the intention to be fictitious \cite{dunne2024speculative}. However, in fashion, such speculative design has been used by designers to consider not only problems the industry faces, such as sustainability \cite{Clarice_2023}, but also personal experiences such as identity and culture \cite{Arora_2023}. While speculative design in fashion has sought to explore more discursive issues, speculative computational wearables\footnote{While Auger–Loizeau's Audio Tooth Implant \cite{Auger2013Spec} could, by some, be considered a wearable, we believe this edges into transhumanism. We see our work maintaining robots and technology as separate entities and are focusing on similar wearable devices.} have mainly looked at how these devices can help with social connection \cite{helen_2019} or help mediate intimacy and touch \cite{Kaur_2023, jewitt_2019}. This suggests that there is still a gap in achieving the type of expression wearers want from their technology, and a need to explore this from a different perspective.

As we have suggested in the previous sections of our related work, soft robotics as wearables could provide new avenues of exploration for computational wearables. However, as Auger's study of living with robots demonstrated, further exploration into how we live with robots is needed \cite{Auger_2014}. Di Silva \cite{DiSalvo_2012} argues that: ``\textit{within the discourses of robotics ... [there are] conflicting claims about what social robots are, could, or should be.}'' Di Silva suggests that engagement with adversarial design could work to address these claims. Specifically, adversarial speculative design aims to provide elements to help spark political debate. We apply elements of adversarial design strategy here to overcome some limitations around the use of prompts and prototypes to spark inquiry in speculative design. In our prototype, we make specific choices, which are not only for aesthetic value, but deliberate design choices to provoke sociotechnical speculation.

\section{The Sumbrella: A Fashionable Soft Robotic Hat}

In this section, we outline the design rationale, fabrication process and system development of our soft robotic technological probe for speculative fashion futures.


\subsection{Design} 

Here we present the Sumbrella (Figure \ref{fig_sumbrella_worn}), a fashionable hat with soft robotic elements that respond to social surroundings through computer vision. We chose the hat as our base for the design of our wearable because proximity to the face, which is prominent for social engagement signalling, was important. The Sumbrella is capable of changing its form to either expose (reveal) or cover (hide) the wearer. It achieves this through controlled inflation patterns that allow one side to lift or descend independently, or for the structure to move in coordinated, symmetrical motions. These transformations are driven by computer vision, which distinguishes between people and objects in the environment. The system is programmed to react positively to the presence of a single individual but to adopt a more defensive configuration when the wearer is approached by multiple people. It also adapts in confined spaces, altering its shape to avoid collision with nearby objects. Control is primarily autonomous, requiring no direct input from the wearer. However, manual input is available: a button press allows the wearer to override the system and select a preset shape. The Sumbrella incorporates kinesic and speculative design elements to spark debate and facilitate exploration of fashion futures.

\subsubsection{Extending Soft Kinesics:}

There is currently limited research into the meaning conveyed by specific soft robotic gestures. As introduced in the related work section, studies have examined directional tilt, inflation and deflation, rate of change, and twisting morphologies \cite{farhadi2022exploring, koike2024sprout}. However, viewer interpretation is far from concordant. Further, leaf-like appendages, morphological transformation resulting from inflation and deflation, and soft robotic waving sequences as communication profiles, have not yet been explored in soft robotic wearables. We decided that innovating actuation would allow us to explore a greater morphological range than simply applying traditional pneumatic systems to our design. Additionally, to our knowledge, no wearable has been developed with the direct intention to convey meaning from a symbiotic human-soft robot collaborative unit. With no concrete design guidelines available, we took an exploratory approach. Taking inspiration from the natural world, we incorporated biomimetic movements into the Sumbrella design.  

When employing biomimicry, we treated the human-robot hybrid as a symbiotic whole. The gestures selected are echoing a wider behavioural paradigm, such as a cat rolling over and stretching, exposing its vulnerable underside to show relaxation and trust \cite{Rault2020}. The Sumbrella sweeps away from the wearer's face to emulate this behaviour, intended to convey a wearer's openness, a welcoming communication. In contrast to this welcoming state, a withdrawal state was developed. This state biomimics defensive strategies that involve puffing out to look bigger, but also hiding away as a form of retreat \cite{edmunds1974defence, Ruxton2007}. Aligning with the finding that a body puffing out is a kinesic soft robotic communication interpreted as defensive \cite{koike2024sprout}. In the withdrawal state, the Sumbrella lowers and stiffens around the wearer's face to form a protective cocoon and communicate a wish for privacy. Finally, kinesic communication in the form of waving sequences were incorporated into Sumbrella's behavioural array as an exploratory element. Previous research has shown that in soft robots, rhythmic, repeating movements can elicit interpretation of an alive being, in a calm and relaxed state \cite{farhadi2022exploring}. However, this is likely an association with breathing, something unlikely to be replicated in appendage embodied motion (as opposed to internal expansion). Through this design choice, we instead sought to explore if rhythmic sequences in the peripheral elements of a soft robot wearable would be more likely to elicit positive interpretations, more akin to the invitation facilitated by a dog wagging its tail \cite{Leonetti2024}. 

\begin{figure}[t]
\includegraphics[width=14.8cm]{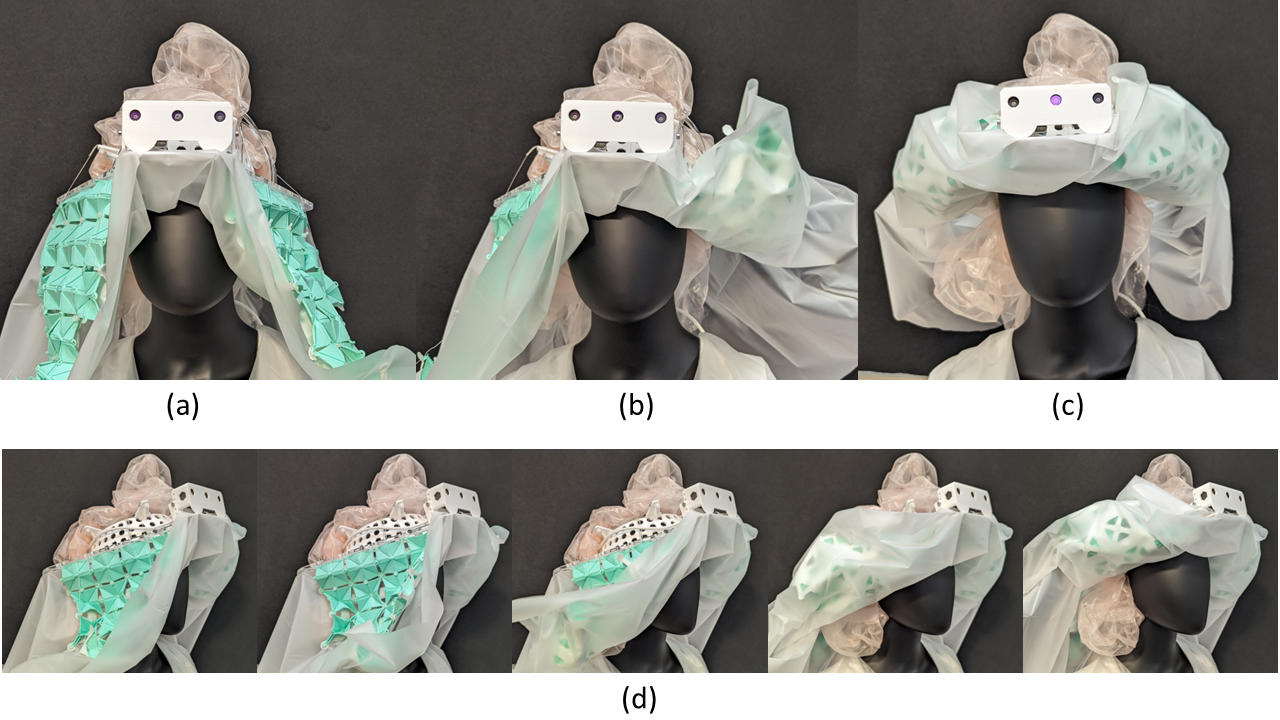}
\caption{Images of the Sumbrella system on a mannequin. (a) Device in neutral withdrawal state with origami modules inflated. Expanded leaves obscure the face and stay stiff to convey a defensive posture. (b) Device raising a leaf showing how it can convey biomimetic gestures via programming, e.g., waving leaves. (c) Device in welcoming state with all leaves raised up, opening up to expose the wearer's face. (d) A sequence of images showing how the leaves raise when not inflated.}
\Description{Photographs of the Sumbrella in various states.}
\label{deviceimage}
\end{figure}

In summary, the behaviours of the system were developed to intuitively emulate welcoming and withdrawal expressions. To welcome a person in and show attention to them, Sumbrella lifts up all its leaves, resembling a normal headpiece with unobstructed views of the wearer's face as shown in Figure \ref{deviceimage} (a-c). When the wearer wants to withdraw and keep to themselves, the leaves are inflated in a released position, puffing up and creating a stiff shroud around the head, akin to a protective shell. Lastly, to catch attention and convey elation, the leaves wave in defined sequences.

\subsubsection{Speculative Elements}

The most distinctly provocative elements of speculative design in Sumbrella's aesthetic composition are the transparent materials employed. It is aesthetically intuitive to hide the functional elements, veiling the garment's construction, thereby tidying the visual effect. In the Sumbrella, we challenge this assumption; everything that can be, has been, purposefully left visible. With the bolero jacket, placing items on the back of the body intentionally showcases the drive and control components and mechanisms. This design strategy not only provides a direct challenge to the assumption that the workings should be hidden, as is common in `finished' objects, but also acts as a material metaphor, harking to the significant area of trust in robotics \cite{hancock2011meta, anjomshoae2019explainable, wortham2017robot}. The transparent textile used throughout evokes a distinct and direct comparison to the conceptual space of transparency in robotic systems. Extending the narratives of trust, more unnerving choices were incorporated. Adversarial design, as noted, is supposed to evoke the political \cite{DiSalvo_2012}. This robot is purposefully designed to make it clear that it watches you. The fabrication methods would permit a hidden sensor-based system, however, a camera is used in the design. Computer vision is a deeply provocative subject \cite{waelen2024ethics}. The use of computer vision here is emphasised by the prominent placement of a distinct pair of eyes, left visible in the casing, resulting in a challengingly obvious, watching robot. This is not only a highly engaging experience for a human \cite{Hamilton2016}, but also an uncanny one, due to the artificiality of the watcher \cite{mori2012uncanny}. Further, the bold gaze of the eyes of the machine, provide a visual clue to the thinking mechanism behind. With artificial intelligence being widely discussed, and little understood \cite{Wei2024, Brauner2023}, it is a timely adversarial, and political element, to design in for provocation of speculative debate.

\subsection{Fabrication \& Motion Analysis}

\begin{figure}[h]
\includegraphics[width=15cm]{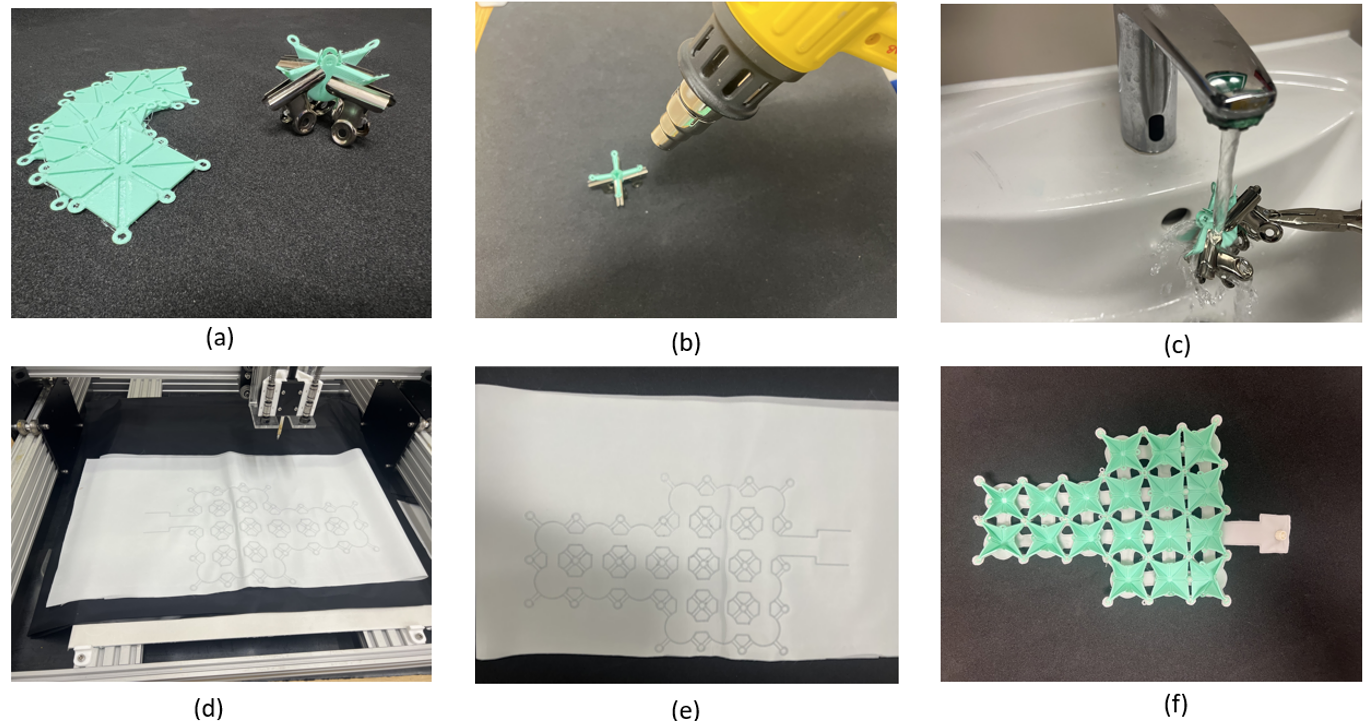}
\caption{Fabrication Process of Sumbrella leaves.}
\Description{Six steps of the fabrication process, (a) to (f), as described in the text.}
\label{FashionPictures}
\end{figure}

In the design of the Sumbrella (as seen in Figure \ref{deviceimage}), there are three leaves. Each of the leaves were made by sewing Thermoplastic Polyurethane (TPU) on white fabric. The designed TPU structures are bistable in nature and take their inspiration from origami. The design was drawn in CAD software and then 3D printed, as shown in Figure \ref{FashionPictures} (a). In the next stage, we need to instill the bistable characteristics into this material. For that, binder clips were used to create the fold as shown in Figure \ref{FashionPictures} (a). Using a heat gun set to 180°C, the material was heated, as shown in Figure \ref{FashionPictures} (b) for about two minutes. It was then cooled by placing it under running water (Figure \ref{FashionPictures} (c)). Once cooled, the clips were removed. The resulting compliant structure exhibited bistability.

A fabric-based pneumatic chamber network was designed in CAD software. The design marks borders where the fabrics will be heat sealed. The 3D CAD model was converted into a 2D plot and which was passed to a computer-controlled 2D heat sealing machine (Figure \ref{FashionPictures} (d)) which controlled the application of a 280°C hot probe to the heat-sealable fabric. The complete fabric network (shown in Figure \ref{FashionPictures} (e)) was then cut from the fabric.

All the bistable TPU structures were sewn onto the fabric. The final design of one of the leaves is shown in the Figure \ref{FashionPictures} (f).

\begin{figure}[h!tbp]
\includegraphics[width=14cm]{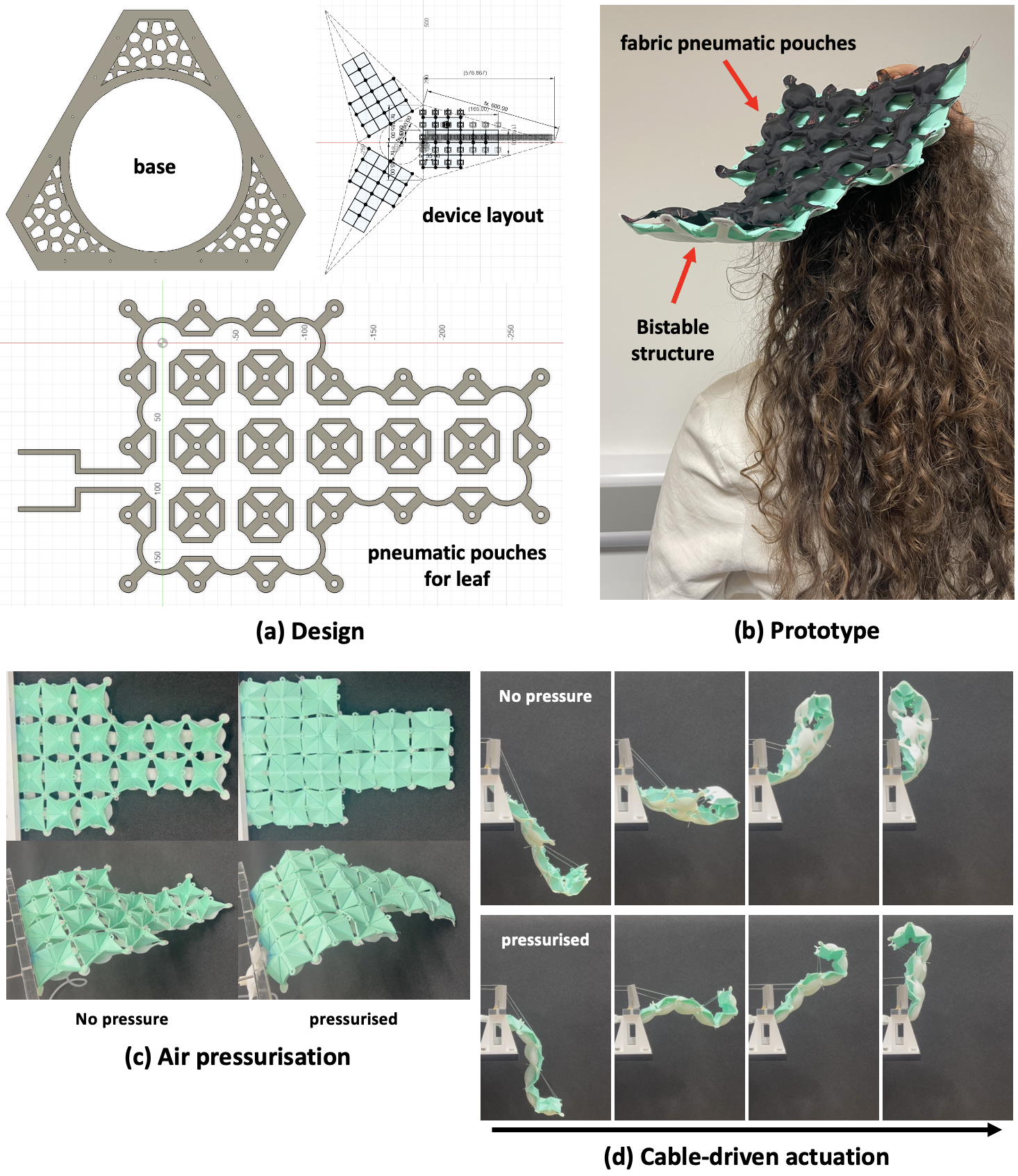}
\caption{The final design elements of the Sumbrella.}
\Description{Design and components of the Sumbrella including (a) the design and layout of the Sumbrella, (b) a look of the inflated leaf worn on a person's head, (c) the pressurisation of the leaf (illustrating the shape change of the soft structural networks) and (d) the cable-driven actuation of the leaf without and with fabric pressurisation (illustrating different motion behaviours).}
\label{Robot}
\end{figure}

To analyse the motion under action, one of the Sumbrella leaves was attached to the end of a test bench as shown in Figure \ref{Robot} (d) and then actuated. Here, we did the experiment when the leaf was not pressurised. The experiment was then repeated with the leaf inflated with air. 

A linear actuator was utilised to pull the leaf through a series of tendons at a constant speed. Figure \ref{Robot} (d) shows the observations made under this study. As could be seen, the actuation behaviour of the leaf varied significantly depending on inflation and cable actuation. Pulling the tendons while uninflated and soft causes the leaf to curl up and retract, while pulling the tendons while inflated and stiff causes the leaf to elevate upwards with far less curling.

\begin{figure}[h]
\includegraphics[width=14.8cm]{Images/Methods/systemandbackview.png}
\caption{System integration of the Sumbrella. Left: an overview diagram of the three subsystems. Right: The back of the system showing how components are placed on the wearable jacket. Inset images above show close-ups of the components in pockets.}
\Description{A diagram of the three subsystems and how they integrate and a photograph showing component placement on the jacket.}
\label{system}
\end{figure}

\subsection{System Integration}

Sumbrella requires pneumatic and motor-driven actuation for each leaf, a way to gather information from the environment, and computing to tie it all together. Figure \ref{system} shows an overview of the different systems and how they are connected, including how they were integrated as part of a jacket for wearability. The system can be broken down into the computing subsystem, the cable drive subsystem, and the pneumatic subsystem. Additional information and code can also be found here: https://github.com/loongyi/Sumbrella. Bill of materials available in supplementary.

To demonstrate the responsive capabilities of the system, a depth-sensing camera (Oak-D Lite) was used to record depth and color (RGBD) information from the environment.  This data was put through a pre-trained MobileNet Single-shot multibox Detection neural network which detects the presence of humans in front of the user. Since the camera can sense depth, the device can estimate the distance of the human from the wearer and preserve personal space. The computing subsystem takes the camera output and feeds it into a Raspberry Pi 4B single board computer, which acts as central processing for the behaviours of the Sumbrella. The computer is programmed with a python script that acts like a finite state machine, with transitions between behaviour states determined by the received camera inputs. These behaviour states contain commands to be sent to the cable drive and pneumatic drive subsystems via serial communications. The states can also be manually changed via a wired controller.

\subsection{System Interaction}

Based on the behaviour states, the cable drive subsystem pulls and releases the cables of the leaves. The cables were routed around the leaf, and fed back to winches on Dynamixel XL 330-M288T servo motors via bowden tubes. The OpenRB-150 microcontroller of the cable drive was also programmed as a devolved state machine which responds to serial commands from the computer. For pull-up behaviours, the motors wind in the cable until a torque threshold is reached, indicating that the leaf has reached its upper limit (motion shown in Figure \ref{deviceimage} (d)). The displacement of the motor is recorded and release behaviours unwind the motor based on this recorded pull-up displacement. For waving behaviours, the leaves are wound up and down with proportional control based on a sine wave. This allows more fluid motions with the ability to change the frequency and phase shift for faster waving and asynchronous waving of the three leaves.

The pneumatic drive system inflates and deflates the leaves to change their bending behaviour. The pneumatic tubes from the leaves are fed into a three way valve which leads to pressure sensors and pumps as illustrated in Figure \ref{system}. The custom STM32-based microcontroller also responds to serial commands from the computer, taking in target chamber pressures and returning the pressure state of each leaf. It was programmed with a Proportional Integral Derivative controller to achieve target pressure inputs by driving the voltage of the air pumps and switching the valves. Each leaf has one pump for inflation, and one pump is used to extract air from all leaves for faster expulsion when the valve is switched to deflate the leaf.

\subsection{Complete Prototype}
To summarise, the Sumbrella is a wearable robotic device that utilises soft robotic technologies to achieve organic actuated motions. These motions were designed to express distinct states (i.e., welcoming and withdrawal) based on the user's input or interaction with the environment. It was designed as a headpiece with three leaves that are laced with origami-inspired bistable popping modules, which pop and change the bending behaviour of the leaves when inflated. Motors with cables retract and release the leaves, rolling them up to unveil the wearers face. The device also has a bolero jacket with pockets which house the electronics and power, enabling autonomous sensing and reaction to people who enter the personal space of the wearer. The transparency of the pockets are not only for aesthetic value, but as a deliberate provocation to encourage critical discourse around robot trust. This work extends some elements of previous work exploring kinesics in soft robotic communication, but is distinct due to the soft robot's relationship with the human body. While the design intends for the robot-human hybrid to be perceived as a whole, there is ambiguity as to whether the meaning should be derived from the soft robot itself, or the wearer. Device capabilities permit programming of a wide range of biomimetic gestures, however, the demonstrations developed are kept simple with respect to the two behaviour modes. 




\section{User Study}

\subsection{Study Design}
We designed our study with two aims in mind: to evaluate the Sumbrella itself in the context of contemporary wear, and to examine speculative futures surrounding robotic augmentation of social processes through fashion. To initiate discussion, we used the prototype as a technological probe \cite{hutchinson2003techprobe}. In a one-hour focus group session, participants were encouraged to handle and interact with the Sumbrella and to engage in a facilitator-led discussion to evaluate the current design and speculatively explore sociotechnical futures \cite{taylor2004modern}. 

\subsection{Participants}
Participants were expert professionals, all residents or alumni of a media studio for creative technologists which focusses on the meeting point of art, technology, and society. These expert professionals apply their knowledge and skills within the studio themes of: interactive documentary, play, location and movement, connected objects, performance and music, storytelling, cities, robotics, moving image, social tech and environmental emergencies. The studio, to foster interdisciplinary collaboration, does not emphasise specific disciplinary backgrounds. However, publicly available resident profiles show the community contributes broad perspectives from across the arts, humanities, sciences and engineering. Our rationale in selecting these participants was to gather diverse perspectives in the evaluation and speculative exploration of the product. 

Participants were recruited through the media studio’s network producer, who promotes a regular session for engaging with early-stage prototypes. Participants were initially informed of the time, date and subject of the session well in advance. Two weeks before the session, participant information sheets providing an outline of the study and describing data capture procedures were circulated. These were shared again two days before the study, and at the media studio immediately prior to the session. Opportunities for research protocol questions and consent were arranged directly prior to the session. Attendance was voluntary and 12 participants attended the session. To respect participant anonymity we did not collect demographic information as the sampling environment is a relatively small organisation and only refer to our participants by `P' plus a numerical (i.e., P1 or P8). However, due to ethical approval processes the age range of participants can be confirmed as between 18 - 65. This is due to the exclusion criteria determining that adults only were recruited for this study, and all participants were active, non-retired, members of a professional community. Based on the studio's published\footnote{We have chosen not to provide a link to this data to, again, protect the privacy of our participants} demographic information, the majority of the residents are between the ages of 35-49 and the studio maintains a general fifty/fifty balance between men and women. 

\subsection{Procedure}
The one-hour study session was conducted as a focus group in the media studio’s event space and structured into three parts: demonstration (15 mins), interaction (20 mins), and discussion (25 mins). The event space was set up with rows of seating facing a main screen and table, these were used for displaying prompt questions and device demonstration. A table set up against the right hand wall was used for the interactive demonstration, and a board provided next to this for sticky notes, which contribute to final discussion. See supplementary materials for a plan of the set up of the event space, for this study (S3).  This session was facilitated by the first author with additional researchers present to provide technical support and explanation, and to take notes. The entire session was audio-recorded.

\textit{Demonstration}\\
The initial focus for the session was on first impressions of the Sumbrella and its suitability as a fashion and social interaction garment. The Sumbrella was placed on a mannequin facing the participants, and a demonstration was conducted. The participants were shown three behaviours from the Sumbrella initiated with manual control via a switch. Participants were shown; a welcoming state, where all leaves were curled up; a withdrawing effect, where the leaves were lowered and stiffened; and an object avoidance state, where single leaves lifted up independently of others in response to human proximity. Following this demonstration, participants were asked for their initial impressions of the Sumbrella. The facilitator asked questions on the subjects of aesthetics and design, functionality, wearability and application. Examples of these questions are ``Is the Sumbrella beautiful, or ugly? Why?'', and ``When the Sumbrella sweeps into its upward shape, what does that communicate to you?''. Additional questions can be found in supplementary.

\textit{Interaction}\\
During this section, participants were given the opportunity to interact with elements of the Sumbrella, and encouraged to start thinking about speculative futures. Three activities were available; handling a single leaf; interacting with the Sumbrella; and responding to speculative prompt questions. Participants self organised their time across these three activities. One spare leaf was passed around; it consisted of one pneumatic fabric pouch with bistable TPU units sewn on and was an exact replica of those in the Sumbrella. Participants were allowed to handle this for as long as they liked to investigate the tactile properties. The Sumbrella was set up at an angle so participants could approach it individually to activate the computer vision system and witness a Sumbrella's reactive behaviour. Participants also had the opportunity to examine and enquire about technical aspects of the device. The program scripts and images being ‘seen’ by the camera were displayed on a monitor for participants to understand the Sumbrella’s function and reactions. While not engaged in either activity, participants individually reflected on questions exploring speculative futures and recorded these on sticky notes. These questions enquired about potential design, social capabilities and relationship with the device. Examples of these questions are ``What could the hat communicate about you?'', and ``What extra responses could the hat have? E.g responded to peoples expressions, working out who you didn’t like, reading who you did like and being nice?''. Refer to supplementary material 1 for additional questions.

\textit{Futures} \\
For the final section, the facilitator led a discussion exploring device-centred speculative futures and wider related sociotechnical imaginaries. This was conducted within a semi-structured approach using the sticky note recorded responses generated in the previous section to prompt discussion. The facilitator grouped sticky notes that had been shared on a board, into similar subjects, then read an example of each subject aloud to invite comment from the group. Engagement from all participants was encouraged through reading of other similar sticky notes, and invitation for the writers to expand on their points. Example questions which prompted reflection on speculative futures were  ``How and where could you wear this in daily life?'' and ``How might public spaces look and feel if many people wore Sumbrellas?''

\subsection{Data Collection and Analytical Framework}
We conducted an inductive reflexive thematic analysis \cite{braun2006thematic} on the transcription using NVivo software \cite{nvivo}. The process began with the first author manually transcribing the audio recording. This manual transcription allowed the first author to familiarise themselves with the data in line with Braun and Clarke's suggested process \cite{braun2006thematic}. In reflexive thematic analysis, understanding the perspectives of the researchers who analysed the data adds value to understanding how their work contributes to knowledge \cite{braun2021one}. The majority of the research team works in soft robotic engineering (except EC, who is a design researcher in human-computer and AI interactions). Specifically, the first author, who conducted the thematic analysis and was the driving force behind this work, is a robopsychologist working at the intersection of robotics and psychology. These interdisciplinary perspectives allowed us to balance and include elements of engineering, design, and psychology. Besides their academic background, the first author has previously worked in the fashion industry for eight years helping situate our work in relation to that industry as well as robotics.

An inductive approach was employed by the first author during coding, as suggested for the open coding phase. At the end of this phase, a total of 112 codes were produced. Three iterations of the coding process were conducted in conjunction with discussions with research team members. During these iterations, axial coding was done to establish relationships between the codes leading to the development of the reoccurring topics of such as \textit{``first impressions-positive'', ``first impressions-negative'', ``interaction with the body'', ``interactions with others'', ``bias''} and \textit{``sense of control''.} These topics were then transferred into Miro\footnote{\url{https://miro.com/}} for the selective coding process, in which the research team further grouped topics to into more nuanced topics: \textit{``first impressions'', ``the body space''} and \textit{``public space''}, (see supplementary material 4). The research team then reevaluated these topics to clarify how their work could create a dialogue with existing computational wearable and soft robotics research. To delve into speculative futures surrounding robotic fashion, we then considered the sociotechnical imaginary \cite{taylor2004modern} as a way of situating the discursive responses to our technological probe. We took guidance from the recent application of the use of sociotechnical imaginary in methodological processes, which consider future digital technology that has not yet been developed or become widespread \cite{Jewitt2020}. Often, the aim of roboticists when developing robots for companionship is to create social agents; however, with Sumbrella, we sought to explore other types of human-robot interaction futures by looking at the wearable robot as a social object \cite{wagner2018making}. By understanding what kind of social object a wearable robot would be, we aimed to look at what behaviours individuals would develop that could allow the robot to serve as a link to their environment, what sort of social relationship they imagined building with the robot, and what collective or interpersonal interactions would arise from having a wearable robot. This resulted in three main themes: \textit{Soft robotic wearable as a positive social object, Personal liabilities that could come from wearing a social object,} and \textit{Dystopian ramifications with `wearing' one's expression.}


\section{Findings}
In our findings, we explore Sumbrella, a speculative wearable soft robot, as a social object. Our participants discussed various positive behaviours and interactions that could arise from owning and wearing something like Sumbrella. However, as can happen with social objects, the participants also mentioned various reservations that they had regarding using Sumbrella in everyday life. These trepidations led our participants to even consider the more dystopian future that could arise through wearable robots as social objects. 

\subsection{Soft robotic wearable as a positive social object}

The first impressions of the Sumbrella from the creative technologists and their imagined impressions from others were relatively positive. Participants felt that the Sumbrella would be a wonderful thing to have at an outdoor music event, with much discussion being focused on the way it would function and the effect it would have in this context. This was based on a playful interpretation of the item, influenced by its aesthetic appearance, behavioural changes and anticipated social effects. Aesthetically, participants felt that it looked like a cross between ocean creatures, ``\textit{it’s like a sea anemone}'' (P8) and period costume ``\textit{it’s giving bonnet}'' (P3); a juxtaposition of forms which could be contributing to humorous interpretations. Addressing RQ2, these findings suggest that participants were able to derive social meaning from the Sumbrella’s novel soft robotic kinesics, with its playful movement and soft biomorphic aesthetics supporting affective interpretations of comfort, humour, and social approachability. Participants particularly felt that at an outdoor music event would be perfect as some drawbacks of the current design, such as the noise of the motors and unpredictability of movement, would not be a problem. The unpredictability of the robot was found to be a fun quality and good for ``\textit{putting it on in the afternoon to show off}'' (P6). 

Additionally, when prompted to imagine that the Sumbrella could read and react based on its wearer's biosignals, i.e., heart or breathing rate, participants expressed how this would help them with interoception -- albeit in a more ambiguous fashion, ``\textit{like a mood ring}'' (P1). In relation to RQ1, this framing reflects a cautious willingness to trust biosignal driven responses, when serving a reflective purpose. Participants felt that having information reflected back to them could be useful, particularly in the context of making one aware of an internal state they had not noticed: ``\textit{I didn’t know that this other bodily thing was kind of happening}'' (P4). Some participants felt that this initial dialogue with Sumbrella could lead to an individual interacting more with the device as they may then use the device to help change their internal state. For instance, if Sumbrella alerted a participant that they were stressed, they may then choose to engage the Sumbrella’s withdrawing capability to create a caring system, ``\textit{`Ok, calm down, we just need a little private space', and you can take a minute and get your heart rate back}'' (P11). Some participants elaborated on this care relationship, further stating that the psychological support could become more conceptual, such as a use case parallel to ``\textit{emotional support water bottles}'' (P8), that the relationship with the object becomes highly elaborated and entangled with emotional welfare. There were some who drew parallels to fidget toys and enjoyed the tactile qualities of, particularly, the sequenced bistable origami units, stating ``\textit{nice to touch}'' (P3) and would be difficult ``\textit{not to play with}'' (P8) if worn. This suggests that Sumbrella could also be a tool to help with emotional regulation. 

Participants noted various positive interactions that Sumbrella could prompt: ``\textit{it would be a hell of an icebreaker}'' (P5). Participants imagine that Sumbrella would help with social communication at events like parties. Participants also imagined how the dynamic movements and shape of Sumbrella could allow for positive collective interactions. For instance, positive aspects of bonding were suggested for an outdoor music event as it would be exceptionally functional as a sunshade, which could balance protection of the wearer with social connection. Again, responding to RQ2, participants noted that Sumbrella’s movements and shape-changing behaviour communicated socially meaningful intent. As a social function, it was suggested that the Sumbrella could not only peel away to make space for others but also extend to socially share its protective capabilities ``\textit{instead of it shutting people out it [could] extend out so someone can come under with you. Like, get under my umbrella}'' (P3). Sumbrella could thus be thought of as a positive social object as it could promote interactions that are beneficial to others.

\subsection{Personal liabilities that could come from wearing a social object} 
While individual interactions with Sumbrella were wholly positive, participants considered that negative ramifications to an individual wearing Sumbrella might occur from others' negative perceptions and interactions. Participants particularly noted that the size and unfamiliar design might result in negative impressions if worn in crowded spaces: ``\textit{you don't want to be the one person on the crowded tube (subway) who's got the huge hat}'' (P7), and observing that in this context ``\textit{you need keep it compact}'' (P6). Participants felt it looked far too conspicuous for daily wear, ``\textit{think about a suit, then there’s an octopus on your head ‘huh, that guy’s got an octopus on his head!’}'' (P6). To solve this, participants felt it should be made sleeker, perhaps by concealing the hat aspect and obvious mechanisms, into a larger garment such as a mackintosh jacket or cape. If this was achieved, it was suggested that it could make its way into daily wear via technology specialist trends like gorpcore \cite{Lee_2019}. The Sumbrella was also considered inappropriate for daily wear due to the unpredictable nature of the behavioural responses. Participants talked about engaging gestural controls and tweaking temporal facets of the activation to achieve a less surprising and smoother result. Finally, there was a gentle awareness of the implications of exclusively user-focused design and the inconvenience it may cause others, ``\textit{often people design these things with the user in mind. But actually, everyone else has to deal with it as well. I’m thinking, Google Glass and so on. Its great if you’re wearing them, but everyone else thinks you’re an absolute dick!}'' (P2). Participants thus described that others' negative perceptions of Sumbrella could cause them to withdraw away from the wearer. 

The negative ramifications to the wearer were further noted when participants considered Sumbrella responding to internal signals of the wearer, directly addressing RQ1 concerning trust in the automation of social intent through computational reading of biosignals. The visible reaction of Sumbrella could result in broadcasting information that the wearer would not want to share. This was particularly evident in the occurrence of personal bodily experiences that had escaped interoceptive awareness, ``\textit{I think there’s medical uses I’d want it to tell me so that no one else knew}'' (P7). This thread of unfocused Sumbrella responses was widely commented on by participants, who stated a lack of trust in the ability of the Sumbrella to interpret their biological signals but display them in a way that matched the social contexts correctly. Many participants felt that social situations were too complex, for example, at a networking event where ``\textit{if it were to purely respond to your heart rate or your levels of anxiety in your body, it would just close up. But actually, that’s the opposite reaction you want to have in that circumstance}'' (P9). This was echoed in examples of social situations which are not relaxing or enjoyable but one would still choose to engage in. Participants also observed that personal preferences surrounding social engagement could be compromised by a biosignal actuated system, not only in terms of revealing private internal state information to others, but also that the Sumbrella may correctly report how one is feeling, which could be problematic: ``\textit{you’d have to be like ‘Oh my, this hat! I don’t know why... it’s not meant to be doing this’}'' (P5).
Sumbrella's potential to reflect the personal body created the dilemma of allowing individuals to have more self-reflection but the concern of these more private reflections being public. Furthermore, when considering how people use fashion to attract or detract attention, participants noted that Sumbrella, ``\textit{might make attracting attention more complex in a public space }'' (P12) as when and how Sumbrella chooses to hide or reveal the wearer to others could cause misunderstandings. As such, Sumbrella, as a social object, would have a complex social relationship with its wearer as it may go against the wearer's desires.   

\subsection{Dystopian ramifications with `wearing' one's expression} 

In considering the wider deployment of complex robotic wearables (RQ3), participants raised a range of social and ethical implications associated with their widespread use. Particularly when thinking about wearing robots in the future, participants contemplated dystopian ways this would impact individuals and society. For example, participants explored how devices like Sumbrella could be vulnerable to misuse by negative actors. Some fears were discussed around the potential for corporations to take over a Sumbrella-type device and use it for profit gains. For example, some participants expressed concerns surrounding the commercialisation of biosignal data, ``\textit{being able to read your mood and then being able to push particular products to you if it relates to your mood.}'' (P10). Others talked about the employment of the device as part of a uniform which could be manipulated to control staff appearances ``\textit{imagine a scary world where the company force says these are the only ... [Sumbrella] shapes you’re allowed to wear when you’re at work}'' (P11) and restrict authentic expression ``\textit{you can imagine like in customer service where someone says, oh you have to wear this, `it shows how open and friendly you are to all the customers'}'' (P11). When interacting with the Sumbrella, one participant’s image lingered on the monitor; observing this, another joked ``\textit{it seems to have taken a photo of you ... you’re on a list now}'' (P2). While jovial, this speaks of deeper narratives surrounding concern about the use of technology to surreptitiously violate privacy or monitor users. 

There was also some consideration of the Sumbrella wearers being susceptible to bias arising from machine learning as it could potentially ``\textit{harden existing biases if it’s learning who you do and don’t like to speak to then ‘I’m going to apply that to all these people’ and then it being a problem}'' (P7). While the incorporation of AI in wearable robots like Sumbrella could help improve the wearer's relationship with the robot (i.e., the application of reinforcement learning could help it match a wearer's preferences \cite{christiano2017deep}) as participants noted, this has implications that could result in wearable robots being seen as a more negative social object. Participants thus worried about how autonomous a wearable robot should be. In response to the example that the Sumbrella could close down if it recognises that you do not want to speak to someone, participants, like P7 stated, ``\textit{that should be my decision.}'' Demonstrating that narratives pushing robots as social agents or individuals is not acceptable to some. 

Finally, participants considered current social inequalities and how future wearable devices like Sumbrella could exacerbate these issues. It was noted that ``\textit{there’s inequality in who can hide themselves in public spaces ... there’s issues there around gender, ethnic identities, which you actually need to be thinking about}'' (P12). This suggests that who could wear Sumbrella and have it be thought of as a positive social object would be limited by social constructs. For instance, ``\textit{some people are more likely to be targeted if they appear to be hiding in a space}'' (P10). Therefore, an object like Sumbrella may only serve as another visual signal that would encourage targeted negative behaviour. The speculative future of inequalities in the connotations around shielding the face was projected into a plausible dystopia: ``\textit{if people know it has a function that shields people's face. At what point do we start having shop owners saying, right I’m banning that from shops because I can’t see your face. It brings up a whole thing, bias, connotations and judgment}'' (P6). Participant concerns over how the Sumbrella could be misused highlighted a range of contemporary issues relating to employment, data and privacy. They also considered gender, identity and ethnicity, urging engagement with social inequalities in the ability to wear certain clothes.  


\section{Discussion}

Our study has highlighted a range of considerations that wearable soft robotic fashion provokes across aesthetics, gesture, emotion, identity, and ethics. Through participant engagement with the Sumbrella, we uncovered how these devices operate not just as tools, but as social objects that mediate complex human concerns. These include questions of control, self-expression, and the cultural implications of wearing robotic garments in public space. The following sections outline key design implications, recommendations and guidelines emerging from these reflections.

\subsection{Implications of Wearable Soft Robots in Society} 
In this section, we discuss the various design implications that were raised by our participants and point out where further design consideration is needed. Particularly highlighted here are the more positive social implications raised in response to RQ3, considering how wearable soft robots such as Sumbrella might support playful, expressive, and socially beneficial forms of interaction when thoughtfully designed.

\subsubsection{Incorporating Positive Design Elements:} The first impression from our participants of the Sumbrella was a playful one. The unpredictable behaviour received a positive interpretation, and in a social context, was considered good for icebreaking. Our participants' positive reaction to the unpredictability of the Sumbrella is a finding rarely discovered in robotic development. We often expect a predictable performance from machines and can find unpredictable robotic movements to be jarring \cite{wozniak}. However, a limited behavioural array does not reflect the richness of interactions normally gained from biological encounters. Interestingly, in the field of socially interactive robots, Alves et al. \cite{Alves_2021} found that unusual characteristics could potentially increase the trustworthiness of the robot. Our findings thus extend Alves' research, by suggesting that the use of an artefact with unpredictable and surprising robot behaviours can elicit fun, playful and positive impressions from participants. 

Furthermore, the wild aesthetics of the Sumbrella cause it to be highly expressive due to the clearly visible mechanisms and large organic shapes priming participants to have an engaging impression \cite{Dagan_2019}. The success of the Sumbrella in producing playful responses could be leveraged further by incorporating more principles from social affordances research into sociotechnical innovation. In particular, designers could consider ways to improve a wearable's shared usage profile \cite{Isbister_2018}. For example, one design request made by one of our participants suggested allowing the canopy to extend out to protect another person as this could be beneficial for creating social connection. We further suggest that such a feature could even be initiated by the wearer, as well as others, to expand into an underexplored area in social wearables which can be activated by the self, and others \cite{Dagan_2019}. 

\noindent\textbf{Summarised Considerations:}
\begin{itemize}
    \item Unpredictability – Can elicit playful and positive impressions when paired with soft form factors and social framing. This suggests that limited unpredictability, when carefully contextualised, may enhance approachability rather than diminish trust. 
    \item Whimsical Aesthetics – Exaggerated organic shapes and visible mechanical components contributed to the Sumbrella’s expressiveness. Making aesthetic choices like this can encourage engagement by resisting traditional expectations of robotic sleekness and control.
    \item Multi-person Interaction – Participant suggestions highlight the potential for shared control and social activation of wearable functions. Designing for co-initiated behaviours may expand the utility of wearables in facilitating social bonding and cooperative use.
\end{itemize}

\subsubsection{Soft Robots with Interoception and Biosignals:} 

In addressing RQ1, this subsection considers issues of trust, control, and interpretability when biosignals and interoceptive data are used to automate social intent in wearable soft robots. The use of wearables to draw attention to an activity which has gone unnoticed through interoception has been widely applied to health contexts \cite{Dunn_2018, keys2025bodydoubt}. In our findings, our participants considered use of an established biofeedback system to activate the Sumbrella. The prospect of using biosignals as an actuation trigger has the potential to create a space for developing sensitivity to oneself \cite{saini2025inflatedata}. This finding echoes the motivation of many previous designers to protect personal space \cite{zheng2017wearable, Buruk2023bodyexten}. However, unlike these previous designers, this design suggestion potentially allows for a way to break away from the traditional proxemics view of space.

Still, using biosignal data to drive robot behaviour is something underexplored. One finding from our research suggests that such difficulty in application stems from the ability of machines to correctly map internally generated signals into correct social responses. Such potential misalignment has been demonstrated in the literature on arousal where it has been shown that approaching strangers on the street can elicit appetitive or aversive responses, which are indeterminate in biosignals  \cite{Engelniedrhammer2019crowding}. To design for this, user control needs to be integrated to allow for the limited ability of machines to interpret high-arousal social settings. For example, when entering a networking situation, which can be highly stressful, settings could be locked to an open state. Or, for a busy public transport location, settings could be set to a protective state. In their research, Toussaint et al. \cite{Toussaint_2020} found that integration of user control in fashion technology has been shown to generate a more positive relationship with garments which use biosignals for actuation, particularly when a person's internal state could be indicated to others. As suggested by our study, to design well for more complex social situations, gestural control integration is also something that should be considered. Gestural control would allow people to seamlessly override automatic responses from their garment dynamically. However, the design of such a system would need to focus on subtle movement to avoid introducing more socially awkward behaviour \cite{Ahlst_2014}.

\noindent\textbf{Summarised Considerations:}
\begin{itemize}
    \item Interoception for Bodily Reflection – Using biosignals as triggers can prompt wearers to attend to internal states that might otherwise remain unnoticed. In wearable robotics, this opens up opportunities beyond health tracking, toward reflective interaction with one’s own emotional or social state, particularly in shared environments.
    \item Balancing Control - Automatic biosignal driven responses may not align with complex social situations. Incorporating user override functions, such as subtle gestural control, helps manage this ambiguity and may improve comfort, agency, and human defined social legibility.
\end{itemize}

\subsubsection{Ethical Implications for autonomous devices in public:}\label{ethical} 

In relation to RQ3, our findings raised a range of social and ethical implications associated with delivering complex robotic wearables for widespread use. These include dystopian realities that can arise if designers do not consider the ethical ramifications of introducing sensing-enabled garments into contexts where consent, privacy, and control may be unevenly distributed. Rushing ahead by incorporating potentially compromising biosignal-integrated garments into the fashion world can, in turn, open up potentially detrimental new avenues of exploitation. For instance, our participants raised concerns around opportunities for exploitation by corporate employers through targeted advertising and concerns over biosignal data misuse. Instead of perpetuating fashion as another means of exploitation, designers need to stay up to date with developments surrounding the ethical use of personal data, and leverage corporate social responsibility strategies \cite{Mathur_2021, holden_2023, Fatima_Elbanna_2022}. 

Another concern brought up by our participants was the use of clothing to create a safe space within a larger public space brings up the question of `hiding'. As an adversarial design choice, Sumbrella's technology (e.g., the camera) were highly visible. However, as Otuu and Sahoo \cite{otuu2025} explored with co-designing wearables for community policing, visibility of technologies is contentious. In contrast to exploring transparency and visibility, fashion design has also explored the development of anti-surveillance fashion, which uses strategies such as patterns and thermally disruptive fabrics to disrupt surveillance technologies \cite{Harvey_2017, Marks_2023}. However, perceptions of people who may wish to wear these clothing options are not equal. For some people, it could be dangerous to wear even a seemingly utilitarian item of clothing. For example, in the case of Black seventeen-year-old Trayvon Martin in the United States, where wearing a hoodie that concealed his face resulted in gun violence and the loss of his life \cite{bell_2013, masi_2017}. Future designers augmenting fashion need to consider differences in perception of people in public spaces, particularly inequalities in freedom to use clothes to hide.  


\noindent\textbf{Summarised Considerations:}
\begin{itemize}
    \item Avoid Exploitation – Designers should critically reflect on how technology may enable new forms of surveillance and control. This includes staying informed on ethical data practices and anticipating misuse in corporate or institutional settings
    \item Design for Unequal Visibility – Designers must consider how garments that afford concealment or protection are perceived differently depending on the wearer's identity. Clothing that enables hiding or retreat may increase safety for some, but risk suspicion or harm for others. Design decisions should account for these unequal freedoms in public space.
\end{itemize}

\subsection{Employment of Soft Robotic Aesthetics and Kinesics in Fashion}
The Sumbrella is a rare example of a soft robotic garment employed as a technological probe to imagine new fashion futures. One aim of this work is to translate what is known in the realm of soft robotic aesthetics and kinesics into the realm of fashion functioning as a social object (in conjunction with the human body and social intention). This work therefore asks whether people understand social meaning from novel soft robotic kinesics when these behaviours are embodied through fashion and worn on the human body (RQ2). The findings here show that the Sumbrella was interpreted as an extension of the human body in certain contexts (such as at a music festival). When the Sumbrella and its wearer were understood as an intentional whole, participants were able to accurately interpret its kinesic gestures, such as openness, as a form of social invitation. This interpretation was made possible through the added function of Sumbrella's appendages, which were designed to elicit and facilitate social interaction through aesthetic expression. Thus, we built on previous findings exploring the human-soft robot hybrid somatic relationship \cite{Belling_2021}, by extending these principles to include the human-soft robot hybrid as a communication symbiote. 

The fairly new concept of soft biomorphism, which engages indirect resemblance of bio-organisms in soft robotic aesthetics, to communicate feelings of well-being and comfort \cite{christiansen2023brings, Christiansen2024ExSilico}, is successfully applied here. Although people translated the abstracted aesthetics of the Sumbrella into biological organisms (e.g., sea creatures), these interpretations were not strictly visual or form-based. Namely, they aligned this evoked sense of familiarity more closely to the Sumbrella’s role as an accompanying object. Participants' descriptions compared more with emotionally supportive artefacts, such as likening it to a treasured water bottle. In addition, participants frequently mentioned comfort and tactility in the context of the Sumbrella’s tangible, physical presence. Tactile interactions elicited by material properties can raise ethical concerns in the context of a worn device (see~\ref{ethical}). However, the replication of the previous findings that soft robotic appendages can prompt tactile exploration \cite{jorgensen2022soft} produced positive outcomes in this context. The success of the Sumbrella in inviting tactile engagement highlights the importance of physical interaction in shaping the overall usage experience. This suggests that the success of soft biomorphism may lie not only in aesthetic cues, but in the overall affective affordances of the wearable.

While it remains unclear whether specific movements of the Sumbrella conveyed precise emotional states, participants understood the device’s overall communicative intent (i.e., when it was welcoming them). Gestures such as the swept-back posture and puffing motions were interpreted as kinesic extensions of the wearer’s intentions. The Sumbrella effectively replicated several outcomes observed in previous soft robotics research, including the expression of abstract biological states through biomorphic forms, the elicitation of comfort, and the invitation to engage through touch. There remains significant work to be done in clarifying how soft robotic kinesics are interpreted, particularly when comparing independently embodied forms with those incorporated into human–robot hybrid systems. Future studies should adopt more structured methodologies to develop clearer design principles and better understand the nuances of this interaction. Regardless, this study extends existing concepts into the emerging space of symbiotic human–robot wearables, demonstrating that soft robotic fashion can operate as more than just a visual or kinetic artefact, it can function as a communicative interface shaped by social and emotional needs.

\noindent\textbf{Design Recommendations:}
\begin{itemize}
    \item Retain and Extend Soft Biomorphism - Soft biomorphism on the body was positively received, reinforcing interpretations linked to bio-organisms and comforting aesthetics. Future designs should retain this quality to support familiar, approachable impressions in wearable robotics.
    \item Maintain the Interaction Profile of Soft Robotic Tangibles - The tactile interaction profile associated with soft robotics remained stable and meaningful in a wearable context. Designers should ensure that these tactile qualities are preserved during integration into garments to maintain engagement.
    \item Support Kinesic Expression through Hybrid Embodiment - When embodied as part of a human–soft robot hybrid, the device enabled clear kinesic expression of social and emotional intent. Designing for hybrid embodiment can enhance the expressive capacity of wearable robotic systems.

\end{itemize}

\subsection{Methodological Reflection on Speculative Design}

Speculative design is a powerful approach for the critical examination of emerging social relationships between humans and wearable soft robots. Unlike traditional design methods, which aim to solve defined problems or optimise performance, speculative design supports inquiry into uncertain futures. More recently, Ringfort-Felner et al. \cite{Ringfort-Felner2025quality} introduced a taxonomy breaking down the qualities of speculative design in HCI into three types of qualities: speculative, discursive, and process. They suggest using their taxonomy to \textit{``facilitate the realization and assessment of speculations.''} 

Ringfort-Felner et al. \cite{Ringfort-Felner2025quality} described speculative qualities found in the literature as being \textit{``fictional, critical, and socio-political.''} Sumbrella was successful in achieving these qualities as it moved beyond being just a technical demonstration of soft robotics by getting participants to imagine a different future with wearable robotics. Sumbrella's visible systems, ambiguity of gestures, and deliberate adversarial cues (i.e., exposed camera components) challenged normative expectations of sleekness, predictability, and control --- leading participants to foreground social and political implications in their discussion. By designing Sumbrella not only to perform, but to provoke, we expanded the role of wearable robotics to include aesthetic exploration and critical foresight through early-stage engagement with social, ethical, and aesthetic questions. 

The next qualities described by Ringfort-Felner et al. \cite{Ringfort-Felner2025quality} in their taxonomy (discursive and process), more simply put, seem to both explore how people engaged with the design. Speculative design, while it has been used in HCI to explore things like transhumanism \cite{Auger_2014}, cultural identity and fashion \cite{Arora_2023}, and AI ethics \cite{hanschke2024} it remains a rare method in soft robotics and engineering-led wearable research, fields in which functionality and technical feasibility still dominate. Thus, unlike Ringfort-Felner et al. \cite{Ringfort-Felner2025quality}, we find it difficult to separate the discursive\footnote{Discursive is separated into Experienceable and Thought-Provoking.} from the process\footnote{Process is separated into Grounded, Participative, Reflected, and Playful.}. In the case of Sumbrella, roboticists contributed complex, high-fidelity systems typically reserved for functional prototypes, to make speculative design experienceable. However, by embedding these systems within a speculative framework and prioritising conceptual provocation over commercial viability, we preserved exploratory potential. This dual approach encouraged engagement with human concerns and reflection early in the production process, without limiting creative pathways through premature optimisation. Fashion has long embraced a similar, layered strategy. Avant-garde collections, often unwearable, serve as cultural provocations that later influence accessible, marketable designs. Likewise, speculative prototypes like Sumbrella can shift discourse, influencing future design directions and value systems in the field of HRI as well as public views on robots. Rather than delivering a product, speculative prototypes seed conversations that will shape how robotics might be worn, understood, and socially accepted. As Forlano notes in reflections on computational fashion, speculative design enables communities to form around contested aesthetics and embedded values, shifting the focus from technical resolution to the lived implications of design futures \cite{forlano2016aesthetics}. Thus, the discursive elements of something being experienceable and thought-provoking should be embedded in the process.

In Sumbrella, this approach supported the creation of a complex, innovative prototype while keeping space open for playful exploration and critical reflection. It helped the team align around shared questions instead of fixed outcomes, and allowed participants to imagine what kinds of futures this technology might invite. For wearable soft robotics, speculative design offers a flexible, repeatable way to surface risks early, spark new ideas, and build technologies that are not only functional but thoughtful, and attuned to the messy realities of human life.

\section{Limitations} 

Due to our process resulting in only one prototype being developed at this stage, there was limited scope for interaction opportunities in our user study. Additional Sumbrellas would give people the opportunity to wear the device, try it in public spaces, and consider social interactions with equal device usage between multiple wearers. The rapid development of our prototype prevented us from sourcing sustainable materials. This was particularly distinct in the search for a commercially available textile which was transparent and waterproof, while also being biodegradable.  With sustainability an essential responsibility in product design, we plan for future prototype iterations to prioritise alternative fabrication methods and consider innovating in the materials space to achieve better outcomes.

\section{Conclusion} 
This research introduces the application of novel soft robotic techniques to fashion as a means to speculatively explore future human-robot interactions. This paper presented insights gained from using our technological artefact as a probe for speculative design. Our thematic analysis revealed a central tension: the same qualities that make soft robotic fashion compelling can also make it ethically and socially fraught. While our participants could imagine Sumbrella as protecting or revealing the wearer positively, they also expressed how this could result in the involuntary sharing or monitoring of one's behaviour. Overall, we found that the soft robotic elements of the Sumbrella successfully leveraged biomimetic kinesics, effectively conveying social meaning and evoking positive associations for participants. This research shows how speculative wearable robotics can materialise complex social questions, making abstract futures tangible and discussable through embodied prototypes. We contribute to HRI reflections on the application of speculative design as a method, and provocations to be considered for developing soft robotics as wearables.

\section{Acknowledgments} 

The authors would like to thank all the participants who joined in with this research, and the media studio, who hosted us for the study. AI was supported by EPSRC grant number EP/Y030974/1. LYL was supported by EP/V062158/1 and EP/V026518/1.
AR was supported by EP/S021795/1.
YZ was supported by  EP/L015293/1.
RSD and NR were supported by EP/S026096/1.
EW was supported by Wellcome Trust Innovator Award 222397/Z/21/Z.
JR was supported through EPSRC research grants EP/V062158/1,
EP/T020792/1, EP/V026518/1, EP/S026096/1 and EP/R02961X/1, and by
the Royal Academy of Engineering as a Chair in Emerging Technologies.

\bibliographystyle{ACM-Reference-Format}
\bibliography{biblio}

@ArtifactSoftware{nvivo,
    title = {NVivo},
    author = {{Lumivero}},
    organization = {Lumivero},
    address = {Denver, CO},
    year = {2023},
    url = {https://lumivero.com/products/nvivo/},
}

@article{Jabari16112024,
author = {Shiva Jabari and Asif Shaikh and Çağlar Genç and Oğuz Buruk and Johanna Virkki and Juho Hamari and},
title = {A Systematic Literature Review on Computational Fashion Wearables},
journal = {International Journal of Human–Computer Interaction},
volume = {40},
number = {22},
pages = {6913--6940},
year = {2024},
publisher = {Taylor \& Francis},
doi = {10.1080/10447318.2023.2269007},

URL = { https://doi.org/10.1080/10447318.2023.2269007},
eprint = { https://doi.org/10.1080/10447318.2023.2269007}
}

@inproceedings{Wang2021Pneuskin,
  author       = {Wang, Y. and Godoy, M.},
  title        = {Pneu-skin: A haptic social interface with inflatable fabrics},
  booktitle    = {Projections - Proceedings of the 26th International Conference of the Association for Computer-Aided Architectural Design Research in Asia, CAADRIA 2021},
  editor       = {Globa, A. and van Ameijde, J. and Fingrut, A. and Kim, N. and Lo, T. T. S.},
  volume       = {1},
  pages        = {713--722},
  year         = {2021},
  publisher    = {The Association for Computer-Aided Architectural Design Research in Asia (CAADRIA)}
}

@inproceedings{Saini2024pneu,
author = {Saini, Aryan and Patibanda, Rakesh and Overdevest, Nathalie and Van Den Hoven, Elise and Mueller, Florian ‘Floyd’},
title = {PneuMa: Designing Pneumatic Bodily Extensions for Supporting Movement in Everyday Life},
year = {2024},
isbn = {9798400704024},
publisher = {Association for Computing Machinery},
address = {New York, NY, USA},
url = {https://doi.org/10.1145/3623509.3633349},
doi = {10.1145/3623509.3633349},
booktitle = {Proceedings of the Eighteenth International Conference on Tangible, Embedded, and Embodied Interaction},
articleno = {1},
numpages = {16},
location = {Cork, Ireland},
series = {TEI '24}
}

@inproceedings{Haynes2024just,
author = {Haynes, Alice C and Kent, Christopher and Rossiter, Jonathan},
title = {Just a Breath Away: Investigating Interactions with and Perceptions of Mediated Breath via a Haptic Cushion},
year = {2024},
isbn = {9798400704024},
publisher = {Association for Computing Machinery},
address = {New York, NY, USA},
url = {https://doi.org/10.1145/3623509.3633384},
doi = {10.1145/3623509.3633384},
booktitle = {Proceedings of the Eighteenth International Conference on Tangible, Embedded, and Embodied Interaction},
articleno = {36},
numpages = {19},
location = {Cork, Ireland},
series = {TEI '24}
}

@inproceedings{Feng2024exbreath,
author = {Feng, Yuan-Ling and Gao, Mingyue and Li, Chih-Heng and Yao, Zhihao and Mi, Haipeng},
title = {ExBreath: Explore the Expressive Breath System as Nonverbal Signs towards Semi-unintentional Expression},
year = {2024},
isbn = {9798400703317},
publisher = {Association for Computing Machinery},
address = {New York, NY, USA},
url = {https://doi.org/10.1145/3613905.3650870},
doi = {10.1145/3613905.3650870},
booktitle = {Extended Abstracts of the CHI Conference on Human Factors in Computing Systems},
articleno = {134},
numpages = {7},
location = {Honolulu, HI, USA},
series = {CHI EA '24}
}

@inproceedings{otuu2025,
author = {Otuu, Obinna Ogbonnia and Sahoo, Deepak},
title = {How Should We Design Technology With Diverse Stakeholders Who Wish Not to Attend Design Activities Together?},
year = {2025},
isbn = {9798400713941},
publisher = {Association for Computing Machinery},
address = {New York, NY, USA},
url = {https://doi.org/10.1145/3706598.3714168},
doi = {10.1145/3706598.3714168},
booktitle = {Proceedings of the 2025 CHI Conference on Human Factors in Computing Systems},
articleno = {486},
numpages = {14},
location = {
},
series = {CHI '25}
}

@inproceedings{Belling_2021,
author = {Belling, Anne-Sofie and Buzzo, Daniel},
title = {The Rhythm of the Robot: A Prolegomenon to Posthuman Somaesthetics},
year = {2021},
isbn = {9781450382137},
publisher = {Association for Computing Machinery},
address = {New York, NY, USA},
url = {https://doi.org/10.1145/3430524.3442470},
doi = {10.1145/3430524.3442470},
booktitle = {Proceedings of the Fifteenth International Conference on Tangible, Embedded, and Embodied Interaction},
articleno = {62},
numpages = {6},
location = {Salzburg, Austria},
series = {TEI '21}
}

@article{wilson2021women,
  title={Women’s bodies and the evolution of anti-rape technologies: From the hoop skirt to the smart frock},
  author={Wilson-Barnao, Caroline and Bevan, Alex and Lincoln, Robyn},
  journal={Body \& Society},
  volume={27},
  number={4},
  pages={30--54},
  year={2021},
  publisher={Sage Publications Sage UK: London, England}
}

@inproceedings{keys2025bodydoubt,
author = {Keys, Rachel and Marshall, Paul and Stuart, Graham and O'Kane, Aisling Ann},
title = {Rethinking Lived Experience in Chronic Illness: Navigating Bodily Doubt with Consumer Technology in Atrial Fibrillation Self-Care},
year = {2025},
isbn = {9798400713941},
publisher = {Association for Computing Machinery},
address = {New York, NY, USA},
url = {https://doi.org/10.1145/3706598.3713326},
doi = {10.1145/3706598.3713326},
booktitle = {Proceedings of the 2025 CHI Conference on Human Factors in Computing Systems},
articleno = {355},
numpages = {19},
location = {
},
series = {CHI '25}
}

@book{dunne2024speculative,
  title={Speculative Everything, With a new preface by the authors: Design, Fiction, and Social Dreaming},
  author={Dunne, Anthony and Raby, Fiona},
  year={2024},
  publisher={MIT press}
}

@article{ometov2021survey,
  title={A survey on wearable technology: History, state-of-the-art and current challenges},
  author={Ometov, Aleksandr and Shubina, Viktoriia and Klus, Lucie and Skibi{\'n}ska, Justyna and Saafi, Salwa and Pascacio, Pavel and Flueratoru, Laura and Gaibor, Darwin Quezada and Chukhno, Nadezhda and Chukhno, Olga and others},
  journal={Computer Networks},
  volume={193},
  pages={108074},
  year={2021},
  publisher={Elsevier}
}

@article{somppi2022dog,
  title={Dog--owner relationship, owner interpretations and dog personality are connected with the emotional reactivity of dogs},
  author={Somppi, Sanni and T{\"o}rnqvist, Heini and Koskela, Aija and Vehkaoja, Antti and Tiira, Katriina and V{\"a}{\"a}t{\"a}j{\"a}, Heli and Surakka, Veikko and Vainio, Outi and Kujala, Miiamaaria V},
  journal={Animals},
  volume={12},
  number={11},
  pages={1338},
  year={2022},
  publisher={MDPI}
}

@article{lefebvre2018swiping,
  title={Swiping me off my feet: Explicating relationship initiation on Tinder},
  author={LeFebvre, Leah E},
  journal={Journal of Social and Personal Relationships},
  volume={35},
  number={9},
  pages={1205--1229},
  year={2018},
  publisher={Sage Publications Sage UK: London, England}
}

@article{Christiansen2024ExSilico,
  author    = {Mads Bering Christiansen and Ahmad Rafsanjani and Jonas J{\o}rgensen},
  title     = {Ex Silico: Soft Biomorphs},
  journal   = {.able Journal},
  year      = {2024},
  url       = {https://able-journal.org/en/ex-silico/}
}

@inproceedings{Devendorf2016i,
author = {Devendorf, Laura and Lo, Joanne and Howell, Noura and Lee, Jung Lin and Gong, Nan-Wei and Karagozler, M. Emre and Fukuhara, Shiho and Poupyrev, Ivan and Paulos, Eric and Ryokai, Kimiko},
title = {"I don't Want to Wear a Screen": Probing Perceptions of and Possibilities for Dynamic Displays on Clothing},
year = {2016},
isbn = {9781450333627},
publisher = {Association for Computing Machinery},
address = {New York, NY, USA},
url = {https://doi.org/10.1145/2858036.2858192},
doi = {10.1145/2858036.2858192},
booktitle = {Proceedings of the 2016 CHI Conference on Human Factors in Computing Systems},
pages = {6028–6039},
numpages = {12},
location = {San Jose, California, USA},
series = {CHI '16}
}

@article{genc2018exploring,
  title={Exploring Computational Materials as Fashion Materials: Recommendations for Designing Fashionable Wearables},
  author={Genç, Çağlar and Buruk, Oğuz Turan and İnal, Sejda and Can, Kemal and Özcan, Oğuzhan},
  journal={International Journal of Design},
  volume={12},
  number={3},
  pages={1--19},
  year={2018},
  url={https://www.ijdesign.org/index.php/IJDesign/article/view/2831}
}

@inproceedings{forlano2016aesthetics,
  author    = {Laura Forlano},
  title     = {Aesthetics, Cosmopolitics and Design Futures in Computational Fashion},
  booktitle = {DRS International Conference 2016: Future-Focused Thinking},
  editor    = {Peter Lloyd and Erik Bohemia},
  pages     = {928--940},
  year      = {2016},
  organization = {Design Research Society},
  address   = {Brighton, United Kingdom},
  month     = {June},
  doi       = {10.21606/drs.2016.373},
  url       = {https://dl.designresearchsociety.org/drs-conference-papers/drs2016/researchpapers/167/}
}

@article{waelen2024ethics,
  author       = {Rosalie Waelen},
  title        = {The ethics of computer vision: an overview in terms of power},
  journal      = {AI and Ethics},
  volume       = {4},
  pages        = {353--362},
  year         = {2024},
  doi          = {10.1007/s43681-023-00272-x},
  publisher    = {Springer},
  issn         = {2730-5953},
  language     = {English}
}

@inproceedings{Buruk2023bodyexten,
author = {Buruk, O\u{g}uz 'Oz' and Matjeka, Louise Petersen and Mueller, Florian ‘Floyd’},
title = {Towards Designing Playful Bodily Extensions: Learning from Expert Interviews},
year = {2023},
isbn = {9781450394215},
publisher = {Association for Computing Machinery},
address = {New York, NY, USA},
url = {https://doi.org/10.1145/3544548.3581165},
doi = {10.1145/3544548.3581165},
booktitle = {Proceedings of the 2023 CHI Conference on Human Factors in Computing Systems},
articleno = {233},
numpages = {20},
location = {Hamburg, Germany},
series = {CHI '23}
}

@article{braun2021one,
  title={One size fits all? What counts as quality practice in (reflexive) thematic analysis?},
  author={Braun, Virginia and Clarke, Victoria},
  journal={Qualitative research in psychology},
  volume={18},
  number={3},
  pages={328--352},
  year={2021},
  publisher={Taylor \& Francis}
}

@article{mori2012uncanny,
  title={The uncanny valley},
  author={Mori, Masahiro and MacDorman, Karl F. and Kageki, Norri},
  journal={IEEE Robotics \& Automation Magazine},
  volume={19},
  number={2},
  pages={98--100},
  year={2012},
  publisher={IEEE}
}

@article{Brauner2023,
  author    = {Philipp Brauner and Alexander Paul Francois Hick and Ralf Philipsen and Martina Ziefle},
  title     = {What does the public think about artificial intelligence?—A criticality map to understand bias in the public perception of AI},
  journal   = {Frontiers in Computer Science},
  volume    = {5},
  year      = {2023},
  doi       = {10.3389/fcomp.2023.1113903},
  url       = {https://www.frontiersin.org/articles/10.3389/fcomp.2023.1113903/full}
}

@article{Wei2024,
  author    = {Mengyi Wei and Yu Feng and Chuan Chen and Peng Luo and Chenyu Zuo and Liqiu Meng},
  title     = {Unveiling public perception of AI ethics: an exploration on Wikipedia data},
  journal   = {EPJ Data Science},
  volume    = {13},
  number    = {1},
  pages     = {26},
  year      = {2024},
  doi       = {10.1140/epjds/s13688-024-00462-5},
  url       = {https://epjdatascience.springeropen.com/articles/10.1140/epjds/s13688-024-00462-5}
}

@article{hancock2011meta,
  author    = {Hancock, Peter A. and Billings, Deborah R. and Schaefer, Kristin E. and Chen, Jessie Y. C. and de Visser, Ewart J. and Parasuraman, Raja},
  title     = {A Meta-Analysis of Factors Affecting Trust in Human-Robot Interaction},
  journal   = {Human Factors},
  volume    = {53},
  number    = {5},
  pages     = {517--527},
  year      = {2011},
  publisher = {SAGE Publications},
  doi       = {10.1177/0018720811417254},
  url       = {https://journals.sagepub.com/doi/10.1177/0018720811417254}
}

@article{Hamilton2016,
  author    = {Antonia F. de C. Hamilton and Fredrik Lind},
  title     = {Audience effects: what can they tell us about social neuroscience, theory of mind and autism?},
  journal   = {Culture and Brain},
  volume    = {4},
  number    = {2},
  pages     = {159--177},
  year      = {2016},
  doi       = {10.1007/s40167-016-0044-5},
  publisher = {Springer}
}

@article{wortham2017robot,
  author = {Wortham, Robert H. and Theodorou, Andreas},
  title = {Robot Transparency, Trust and Utility},
  journal = {Connection Science},
  volume = {29},
  number = {3},
  pages = {242--248},
  year = {2017},
  publisher = {Taylor & Francis},
  doi = {10.1080/09540091.2017.1313816},
  url = {https://doi.org/10.1080/09540091.2017.1313816}
}

@inproceedings{anjomshoae2019explainable,
  author = {Anjomshoae, Sule and Najjar, Amro and Calvaresi, Davide and Fr\"{a}mling, Kary},
  title = {Explainable Agents and Robots: Results from a Systematic Literature Review},
  booktitle = {Proceedings of the 18th International Conference on Autonomous Agents and MultiAgent Systems},
  year = {2019},
  isbn = {9781450363099},
  publisher = {International Foundation for Autonomous Agents and Multiagent Systems},
  address = {Richland, SC},
  pages = {1078--1088},
  numpages = {11},
  location = {Montreal QC, Canada},
  series = {AAMAS '19}
}

@inproceedings{saini2025inflatedata,
author = {Saini, Aryan and Vs, Sabari and Montoya, Maria Fernanda and Overdevest, Nathalie and Patibanda, Rakesh and Elvitigala, Don Samitha and van den Hoven, Elise and Mueller, Florian ‘Floyd’},
title = {Inflated Exertion: Designing a Bodily Extension that Embodies Physical Activity},
year = {2025},
isbn = {9798400711978},
publisher = {Association for Computing Machinery},
address = {New York, NY, USA},
url = {https://doi.org/10.1145/3689050.3706069},
doi = {10.1145/3689050.3706069},
booktitle = {Proceedings of the Nineteenth International Conference on Tangible, Embedded, and Embodied Interaction},
articleno = {76},
numpages = {7},
location = {
},
series = {TEI '25}
}

@article{Rault2020,
  author    = {Jean-Loup Rault and Susanne Waiblinger and Xavier Boivin and Paul Hemsworth},
  title     = {The Power of a Positive Human--Animal Relationship for Animal Welfare},
  journal   = {Frontiers in Veterinary Science},
  volume    = {7},
  pages     = {590867},
  year      = {2020},
  publisher = {Frontiers Media S.A.},
  doi       = {10.3389/fvets.2020.590867},
  url       = {https://doi.org/10.3389/fvets.2020.590867}
}

@inproceedings{hutchinson2003techprobe,
author = {Hutchinson, Hilary and Mackay, Wendy and Westerlund, Bo and Bederson, Benjamin B. and Druin, Allison and Plaisant, Catherine and Beaudouin-Lafon, Michel and Conversy, St\'{e}phane and Evans, Helen and Hansen, Heiko and Roussel, Nicolas and Eiderb\"{a}ck, Bj\"{o}rn},
title = {Technology probes: inspiring design for and with families},
year = {2003},
isbn = {1581136307},
publisher = {Association for Computing Machinery},
address = {New York, NY, USA},
url = {https://doi.org/10.1145/642611.642616},
doi = {10.1145/642611.642616},
booktitle = {Proceedings of the SIGCHI Conference on Human Factors in Computing Systems},
pages = {17–24},
numpages = {8},
location = {Ft. Lauderdale, Florida, USA},
series = {CHI '03}
}

@article{hrga2019wearable,
  title={Wearable Technologies: Between Fashion, Art, Performance, and Science (Fiction).},
  author={Hrga, Iztok},
  journal={Tekstilec},
  volume={62},
  number={2},
  year={2019}
}

@inproceedings{zheng2017wearable,
  title={Wearable aura: an interactive projection on personal space to enhance communication},
  author={Zheng, Dingding and Lugaresi, Laura and Chernyshov, George and Tag, Benjamin and Inakage, Masa and Kunze, Kai},
  booktitle={Proceedings of the 2017 ACM International Joint Conference on Pervasive and Ubiquitous Computing and Proceedings of the 2017 ACM International Symposium on Wearable Computers},
  pages={141--144},
  year={2017}
}

@article{Engelniedrhammer2019crowding,
author = {Engelniederhammer, Anna and Papastefanou, Georgios  and Xiang, Luyao},
title = {Crowding density in urban environment and its effects on emotional responding of pedestrians: Using wearable device technology with sensors capturing proximity and psychophysiological emotion responses while walking in the street},
journal = {Journal of Human Behavior in the Social Environment},
volume = {29},
number = {5},
pages = {630-646},
year = {2019},
publisher = {Routledge},
doi = {10.1080/10911359.2019.1579149}
}

@book{taylor2004modern,
  title={Modern social imaginaries},
  author={Taylor, Charles},
  year={2004},
  publisher={Duke University Press}
}

@Inbook{Jewitt2020,
author="Jewitt, Carey
and Price, Sara
and Leder Mackley, Kerstin
and Yiannoutsou, Nikoleta
and Atkinson, Douglas",
title="Sociotechnical Imaginaries of Digital Touch",
bookTitle="Interdisciplinary Insights for Digital Touch Communication",
year="2020",
publisher="Springer International Publishing",
address="Cham",
pages="89--106",
isbn="978-3-030-24564-1",
doi="10.1007/978-3-030-24564-1_6",
url="https://doi.org/10.1007/978-3-030-24564-1_6"
}

@article{braun2006thematic,
author = {Braun, Virginia and Clarke, Victoria},
year = {2006},
month = {01},
pages = {77-101},
title = {Using thematic analysis in psychology},
volume = {3},
journal = {Qualitative Research in Psychology},
doi = {10.1191/1478088706qp063oa}
}

@inproceedings{schaar2021predicting,
  title={Predicting Inclusive Futures: Wearables, Automation, and Design Speculation},
  author={Schaar, Raja and Zeagler, Clint},
  booktitle={Advances in Industrial Design: Proceedings of the AHFE 2021 Virtual Conferences on Design for Inclusion, Affective and Pleasurable Design, Interdisciplinary Practice in Industrial Design, Kansei Engineering, and Human Factors for Apparel and Textile Engineering, July 25-29, 2021, USA},
  pages={157--164},
  year={2021},
  organization={Springer}
}

@online{SmokeDress2012,
    author = {Wipprecht, Anouk},
    title = {Smoke Dress},
    year = {2012},
    url = {http://streamingmuseum.org/anouk-wipprechtwhat-does-fashion-lack-microcontrollers/},
    note = {Accessed 31 January 2024}
}

@misc{Multiply_2020, 
title={Social distance in style with Multiply’s “Petticoat dress”}, 
url={https://www.designboom.com/design/social-distance-style-with-multiply-petticoat-dress-05-22-2020/}, 
journal={designboom}, 
author={Multiply}, 
year={2020}, 
month={Jun},
note = {Accessed 01 February 2024}
}

@misc{Quah_2015, 
title={Expanding dress protects the wearer’s personal space}, 
url={https://www.dezeen.com/2014/05/31/diy-dress-expands-to-protect-womens-personal-space-kathleen-mcdermott/}, 
journal={Dezeen}, 
author={Quah, Grace}, 
year={2015}, 
month={Oct},
note = {Accessed 01 February 2024}
}

@misc{Intravaia_2020, 
url={https://boudoirnumerique.com/magazine-en/proximity-dress-social-distancing-revisited-by-anouk-wipprechts-electronic-couture-59123}, 
journal={Le boudoir numérique}, 
publisher={Le boudoir numérique}, 
author={Intravaia, Ludmilla}, 
year={2020}, 
month={Sep},
note = {Accessed 01 February 2024}
}

@inproceedings{harrington,
author = {Harrington, Christina and Dillahunt, Tawanna R},
title = {Eliciting Tech Futures Among Black Young Adults: A Case Study of Remote Speculative Co-Design},
year = {2021},
isbn = {9781450380966},
publisher = {Association for Computing Machinery},
address = {New York, NY, USA},
url = {https://doi.org/10.1145/3411764.3445723},
doi = {10.1145/3411764.3445723},
booktitle = {Proceedings of the 2021 CHI Conference on Human Factors in Computing Systems},
articleno = {397},
numpages = {15},
location = {Yokohama, Japan},
series = {CHI '21}
}

@inproceedings{wozniak,
author = {Wo\'{z}niak, Pawe\l{} W. and Karolus, Jakob and Lang, Florian and Eckerth, Caroline and Sch\"{o}ning, Johannes and Rogers, Yvonne and Niess, Jasmin},
title = {Creepy Technology:What Is It and How Do You Measure It?},
year = {2021},
isbn = {9781450380966},
publisher = {Association for Computing Machinery},
address = {New York, NY, USA},
url = {https://doi.org/10.1145/3411764.3445299},
doi = {10.1145/3411764.3445299},
booktitle = {Proceedings of the 2021 CHI Conference on Human Factors in Computing Systems},
articleno = {719},
numpages = {13},
location = {Yokohama, Japan},
series = {CHI '21}
}

@misc{Art_2021, 
title={Feyfey Yufei Liu}, 
url={https://2021.rca.ac.uk/students/yufei-liu/}, 
journal={RCA 2021}, author={Art, Royal College of}, 
year={2021}, 
month={Oct},
note = {Accessed 06 Februaury 2024}
}

@misc{Levy_2017, 
title={Ying Gao’s dresses become animated “in the presence of strangers”}, 
url={https://www.dezeen.com/2017/10/22/ying-gaos-dresses-become-animated-in-the-presence-of-strangers/}, 
journal={Dezeen}, 
author={Levy, Natasha}, 
year={2017}, 
month={Oct},
note = {Accessed 06 Februaury 2024}
}

@article{edmunds1974defence,
  title={Defence in animals: a survey of anti-predator defences},
  author={Edmunds, Malcolm},
  journal={(No Title)},
  year={1974}
}

@article{Ruxton2007,
  author       = {Graeme D. Ruxton and Thomas N. Sherratt and Michael P. Speed},
  title        = {Avoiding attack: the evolutionary ecology of crypsis, warning signals and mimicry},
  journal      = {ScienceDirect},
  year         = {2007},
  volume       = {8},
  number       = {3},
  pages        = {186--195},
  doi          = {10.1016/j.tree.2007.01.005},
  url          = {https://www.sciencedirect.com/science/article/pii/S1569733907000057}
}

@article{Leonetti2024,
  author = {Leonetti, S. and Cimarelli, G. and Hersh, T. A. and Ravignani, A.},
  title = {Why do dogs wag their tails?},
  journal = {Biology Letters},
  volume = {20},
  number = {1},
  pages = {20230407},
  year = {2024},
  doi = {10.1098/rsbl.2023.0407},
  url = {https://royalsocietypublishing.org/doi/10.1098/rsbl.2023.0407}
}

@misc{Howarth_2014, 
title={Gaze-activated dresses with eye-tracking technology by Ying Gao}, 
url={https://www.dezeen.com/2013/06/24/nowhere-nowhere-two-gaze-activated-glow-in-the-dark-dresses-eye-tracking-ying-gao/}, 
journal={Dezeen}, 
author={Howarth, Dan}, 
year={2014}, 
month={Feb},
note = {Accessed 31 January 2024}
}

@misc{Gonzalez_2016, 
title={3-D printed garment shape-shifts based on an onlooker’s gaze}, 
url={https://www.wired.com/2016/01/3-d-printed-garment-shape-shifts-based-on-an-onlookers-gaze/}, 
journal={Wired}, 
publisher={Conde Nast}, 
author={Gonzalez, Robbie}, 
year={2016}, 
month={Jan},
note = {Accessed 31 January 2024}
}

@misc{jeong_2013, 
title={Trans-for-m-otion}, 
url={https://eunjeongjeon.wixsite.com/infor/projectpage-cv7b}, 
journal={infor}, 
author={jeong, eunjeong},
year={2013},
note = {Accessed 31 January 2024}
}

@inproceedings{Alves_2021,
author = {Alves-Oliveira, Patr\'{\i}cia and Lupetti, Maria Luce and Luria, Michal and L\"{o}ffler, Diana and Gamboa, Mafalda and Albaugh, Lea and Kamino, Waki and K. Ostrowski, Anastasia and Puljiz, David and Reynolds-Cu\'{e}llar, Pedro and Scheunemann, Marcus and Suguitan, Michael and Lockton, Dan},
title = {Collection of Metaphors for Human-Robot Interaction},
year = {2021},
isbn = {9781450384766},
publisher = {Association for Computing Machinery},
address = {New York, NY, USA},
url = {https://doi.org/10.1145/3461778.3462060},
doi = {10.1145/3461778.3462060},
booktitle = {Proceedings of the 2021 ACM Designing Interactive Systems Conference},
pages = {1366–1379},
numpages = {14},
location = {Virtual Event, USA},
series = {DIS '21}
}

@inproceedings{Dagan_2019,
author = {Dagan, Ella and M\'{a}rquez Segura, Elena and Altarriba Bertran, Ferran and Flores, Miguel and Mitchell, Robb and Isbister, Katherine},
title = {Design Framework for Social Wearables},
year = {2019},
isbn = {9781450358507},
publisher = {Association for Computing Machinery},
address = {New York, NY, USA},
url = {https://doi.org/10.1145/3322276.3322291},
doi = {10.1145/3322276.3322291},
booktitle = {Proceedings of the 2019 on Designing Interactive Systems Conference},
pages = {1001–1015},
numpages = {15},
location = {San Diego, CA, USA},
series = {DIS '19}
}

@INBOOK{DiSalvo_2012,
author={DiSalvo, Carl},
booktitle={Adversarial Design}, 
title={Adversarial Design as Inquiry and Practice}, 
year={2012},
volume={},
number={},
pages={115-125},
doi={}
}

@article{Dunn_2018,
author = {Dunn, Jessilyn and Runge, Ryan and Snyder, Michael},
title = {Wearables and the medical revolution},
journal = {Personalized Medicine},
volume = {15},
number = {5},
pages = {429-448},
year = {2018},
doi = {10.2217/pme-2018-0044},
    note ={PMID: 30259801},
URL = {https://doi.org/10.2217/pme-2018-0044},
eprint = { https://doi.org/10.2217/pme-2018-0044}
}

@inproceedings{Toussaint_2020,
author = {Toussaint, Lianne and Toeters, Marina},
title = {Keeping the Data within the Garment: Balancing Sensing and Actuating in Fashion Technology},
year = {2020},
isbn = {9781450369749},
publisher = {Association for Computing Machinery},
address = {New York, NY, USA},
url = {https://doi.org/10.1145/3357236.3395577},
doi = {10.1145/3357236.3395577},
booktitle = {Proceedings of the 2020 ACM Designing Interactive Systems Conference},
pages = {2229–2238},
numpages = {10},
location = {Eindhoven, Netherlands},
series = {DIS '20}
}

@inproceedings{Ahlst_2014,
author = {Ahlstr\"{o}m, David and Hasan, Khalad and Irani, Pourang},
title = {Are you comfortable doing that? acceptance studies of around-device gestures in and for public settings},
year = {2014},
isbn = {9781450330046},
publisher = {Association for Computing Machinery},
address = {New York, NY, USA},
url = {https://doi.org/10.1145/2628363.2628381},
doi = {10.1145/2628363.2628381},
booktitle = {Proceedings of the 16th International Conference on Human-Computer Interaction with Mobile Devices \& Services},
pages = {193–202},
numpages = {10},
location = {Toronto, ON, Canada},
series = {MobileHCI '14}
}

@inproceedings{Isbister_2018,
author = {Isbister, Katherine and M\'{a}rquez Segura, Elena and Melcer, Edward F.},
title = {Social Affordances at Play: Game Design Toward Socio-Technical Innovation},
year = {2018},
isbn = {9781450356206},
publisher = {Association for Computing Machinery},
address = {New York, NY, USA},
url = {https://doi.org/10.1145/3173574.3173946},
doi = {10.1145/3173574.3173946},
booktitle = {Proceedings of the 2018 CHI Conference on Human Factors in Computing Systems},
pages = {1–10},
numpages = {10},
location = {Montreal QC, Canada},
series = {CHI '18}
}

@misc{Marks_2023, 
title={Cap\_able blocks facial recognition software with knitted clothing}, 
url={https://www.dezeen.com/2023/02/07/cap_able-facial-recognition-blocking-clothing/?li_source=LI&li_medium=rhs_block_1}, 
journal={Dezeen}, 
author={Marks, Anna},
year={2023}, 
month={Feb},
note = {Accessed 31 January 2024}
}

@misc{Harvey_2017, 
title={Stealth wear}, 
url={https://adam.harvey.studio/stealth-wear/}, 
journal={Stealth Wear - Adam Harvey Studio}, 
author={Harvey, Adam}, 
year={2017}, 
month={Apr},
note = {Accessed 31 January 2024}
}

@article{bell_2013,
author = {Charlotte Bell},
title = {The Inner City and the ‘Hoodie’},
journal = {Wasafiri},
volume = {28},
number = {4},
pages = {38-44},
year = {2013},
publisher = {Routledge},
doi = {10.1080/02690055.2013.826885},
URL = {https://doi.org/10.1080/02690055.2013.826885},
eprint = {https://doi.org/10.1080/02690055.2013.826885}
}

@article{masi_2017,
author = {Erynn Masi de Casanova and Curtis L. Webb, III},
title ={A Tale of Two Hoodies},
journal = {Men and Masculinities},
volume = {20},
number = {1},
pages = {117-122},
year = {2017},
doi = {10.1177/1097184X17696363},
URL = {https://doi.org/10.1177/1097184X17696363},
eprint = {https://doi.org/10.1177/1097184X17696363}
}

@inproceedings{Mathur_2021,
author = {Mathur, Arunesh and Kshirsagar, Mihir and Mayer, Jonathan},
title = {What Makes a Dark Pattern... Dark? Design Attributes, Normative Considerations, and Measurement Methods},
year = {2021},
isbn = {9781450380966},
publisher = {Association for Computing Machinery},
address = {New York, NY, USA},
url = {https://doi.org/10.1145/3411764.3445610},
doi = {10.1145/3411764.3445610},
booktitle = {Proceedings of the 2021 CHI Conference on Human Factors in Computing Systems},
articleno = {360},
numpages = {18},
location = {Yokohama, Japan},
series = {CHI '21}
}

@InProceedings{holden_2023,
author="Holden, Ben
and Duffy, Vincent G.",
editor="Degen, Helmut
and Ntoa, Stavroula
and Moallem, Abbas",
title="Wearable Technology Safety and Ethics: A Systematic Review and Reappraisal",
booktitle="HCI International 2023 -- Late Breaking Papers",
year="2023",
publisher="Springer Nature Switzerland",
address="Cham",
pages="390--407",
isbn="978-3-031-48057-7"
}

@article{Fatima_Elbanna_2022, 
title={Corporate Social Responsibility (CSR) implementation: A Review and a research agenda towards an integrative framework}, 
volume={183}, 
DOI={10.1007/s10551-022-05047-8}, 
number={1}, 
journal={Journal of Business Ethics}, 
author={Fatima, Tahniyath and Elbanna, Said}, 
year={2022}, 
month={Feb}, 
pages={105–121}
}

@article{Lee_2019, 
title={Semantic network analysis of Gorpcore},
volume={76}, 
DOI={10.31274/itaa.8218}, 
number={1}, 
journal={International Textile and Apparel Association Annual Conference Proceedings}, 
author={Lee, Hyun-Jung and Shin, HyunJu and Lee, Kyu-Hye and Lee, Seulah and In, Ye-Jin}, 
year={2019}, 
month={Jan}
}

@inbook{Galloway_2018,
author = {Galloway, Anne and Caudwell, Catherine},
year = {2018},
month = {10},
pages = {85-96},
title = {Speculative design as research method: From answers to questions and “staying with the trouble”},
isbn = {9781315526379},
journal = {Undesign: Critical Practices at the Intersection of Art and Design},
doi = {10.4324/9781315526379-8}
}

@Inbook{GamitoCaixas2023,
author="Gamito Caixas, Maria Jo{\~a}o
and Matias Sanches Oliveira, Fernando Jorge",
editor="Martins, Nuno
and Raposo, Daniel",
title="Speculative for Strategic Design",
bookTitle="Communication Design and Branding: A Multidisciplinary Approach",
year="2023",
publisher="Springer Nature Switzerland",
address="Cham",
pages="295--304",
isbn="978-3-031-35385-7",
doi="10.1007/978-3-031-35385-7_17",
url="https://doi.org/10.1007/978-3-031-35385-7_17"
}

@article{Clarice_2023,
title = {Fashion futuring: Intertwining speculative design, foresight and material culture towards sustainable futures},
journal = {Futures},
volume = {153},
pages = {103242},
year = {2023},
issn = {0016-3287},
doi = {https://doi.org/10.1016/j.futures.2023.103242},
url = {https://www.sciencedirect.com/science/article/pii/S0016328723001465},
author = {Clarice {Carvalho Garcia}}
}

@misc{alexander2011,
	title = {Getting {Political}},
	url = {https://www.vogue.co.uk/article/lady-gaga-meat-dress-explanation},
	language = {en-GB},
	urldate = {2025-05-23},
	journal = {British Vogue},
	author = {Alexander, Ella},
	month = jun,
	year = {2011},
	note = {Section: tags},
}

@misc{ostrich,
	title = {Original {Napping} {Pillow} – {Ostrichpillow}},
author = {Ostrichpillow},
	url = {https://ostrichpillow.co.uk/products/original-napping-pillow},
	urldate = {2025-05-23},
}

@inproceedings{Arora_2023,
author = {Arora, Sarthak and Kaur, Sachleen and Kar, Ritwik and Sharma, Sumita and Eden, Grace},
title = {What Not to Wear: Exploring Taboos in Clothing Through Speculative Design},
year = {2023},
isbn = {9781450398930},
publisher = {Association for Computing Machinery},
address = {New York, NY, USA},
url = {https://doi.org/10.1145/3563657.3595972},
doi = {10.1145/3563657.3595972},
booktitle = {Proceedings of the 2023 ACM Designing Interactive Systems Conference},
pages = {1–14},
numpages = {14},
location = {Pittsburgh, PA, USA},
series = {DIS '23}
}

@article{forlano_2014,
author = {Laura Forlano and Anijo Mathew},
title = {From Design Fiction to Design Friction: Speculative and Participatory Design of Values-Embedded Urban Technology},
journal = {Journal of Urban Technology},
volume = {21},
number = {4},
pages = {7-24},
year = {2014},
publisher = {Routledge},
doi = {10.1080/10630732.2014.971525},
URL = {https://doi.org/10.1080/10630732.2014.971525},
eprint = {https://doi.org/10.1080/10630732.2014.971525}
}

@inproceedings{stals_2019,
author = {Stals, Shenando and Smyth, Michael and Mival, Oli},
title = {UrbanIxD: From Ethnography to Speculative Design Fictions for the Hybrid City},
year = {2019},
isbn = {9781450372039},
publisher = {Association for Computing Machinery},
address = {New York, NY, USA},
url = {https://doi.org/10.1145/3363384.3363486},
doi = {10.1145/3363384.3363486},
booktitle = {Proceedings of the Halfway to the Future Symposium 2019},
articleno = {42},
numpages = {10},
location = {Nottingham, United Kingdom},
series = {HTTF 2019}
}

@inproceedings{Ringfort-Felner2025quality,
author = {Ringfort-Felner, Ronda and D\"{o}rrenb\"{a}cher, Judith and Hassenzahl, Marc},
title = {The Quality of Speculation – A Scoping Review},
year = {2025},
isbn = {9798400714856},
publisher = {Association for Computing Machinery},
address = {New York, NY, USA},
url = {https://doi.org/10.1145/3715336.3735794},
doi = {10.1145/3715336.3735794},
booktitle = {Proceedings of the 2025 ACM Designing Interactive Systems Conference},
pages = {2373–2394},
numpages = {22},
location = {
},
series = {DIS '25}
}

@article{Auger_2014,
author = {Auger, James},
title = {Living with robots: a speculative design approach},
year = {2014},
issue_date = {February 2014},
publisher = {Journal of Human-Robot Interaction Steering Committee},
volume = {3},
number = {1},
url = {https://doi.org/10.5898/JHRI.3.1.Auger},
doi = {10.5898/JHRI.3.1.Auger},
journal = {J. Hum.-Robot Interact.},
month = {feb},
pages = {20–42},
numpages = {23}
}

@inproceedings{hanschke2024,
author = {Hanschke, Vanessa Aisyahsari and Rees, Dylan and Alanyali, Merve and Hopkinson, David and Marshall, Paul},
title = {Data Ethics Emergency Drill: A Toolbox for Discussing Responsible AI for Industry Teams},
year = {2024},
isbn = {9798400703300},
publisher = {Association for Computing Machinery},
address = {New York, NY, USA},
url = {https://doi.org/10.1145/3613904.3642402},
doi = {10.1145/3613904.3642402},
booktitle = {Proceedings of the 2024 CHI Conference on Human Factors in Computing Systems},
articleno = {470},
numpages = {17},
location = {Honolulu, HI, USA},
series = {CHI '24}
}

@inproceedings{gen_2023,
author = {Gen\c{c}, \c{C}a\u{g}lar and Spors, Velvet and Buruk, O\u{g}uz Oz and Thibault, Mattia and Masek, Leland and Hamari, Juho},
title = {Crafting Bodies and Auras: Speculative Designs for Transhuman Communication},
year = {2023},
isbn = {9798400708749},
publisher = {Association for Computing Machinery},
address = {New York, NY, USA},
url = {https://doi.org/10.1145/3616961.3617810},
doi = {10.1145/3616961.3617810},
booktitle = {Proceedings of the 26th International Academic Mindtrek Conference},
pages = {354–358},
numpages = {5},
location = {Tampere, Finland},
series = {Mindtrek '23}
}

@inproceedings{Kaur_2023,
author = {Kaur, Jasmine and Devgon, Rishabh and Goel, Smera and Singh, Arunesh and Monteiro, Kyzyl and Singh, Advika},
title = {Future of Intimate Artefacts: A Speculative Design Investigation},
year = {2023},
isbn = {9781450399821},
publisher = {Association for Computing Machinery},
address = {New York, NY, USA},
url = {https://doi.org/10.1145/3570211.3570216},
doi = {10.1145/3570211.3570216},
booktitle = {Proceedings of the 13th Indian Conference on Human-Computer Interaction},
pages = {57–66},
numpages = {10},
location = {Hyderabad, India},
series = {IndiaHCI '22}
}

@article{Auger2013Spec,
author = {James Auger and},
title = {Speculative design: crafting the speculation},
journal = {Digital Creativity},
volume = {24},
number = {1},
pages = {11--35},
year = {2013},
publisher = {CAA Website},
doi = {10.1080/14626268.2013.767276},
URL = { https://doi.org/10.1080/14626268.2013.767276},
eprint = { https://doi.org/10.1080/14626268.2013.767276}
}

@article{jewitt_2019,
author = {Carey Jewitt and Kerstin Leder Mackley and Sara Price},
title ={Digital touch for remote personal communication: An emergent sociotechnical imaginary},
journal = {New Media \& Society},
volume = {23},
number = {1},
pages = {99-120},
year = {2021},
doi = {10.1177/1461444819894304},
URL = {https://doi.org/10.1177/1461444819894304},
eprint = {https://doi.org/10.1177/1461444819894304}
}

@inproceedings{helen_2019,
author = {Oliver, Helen},
title = {Design Fiction for Real-World Connected Wearables},
year = {2019},
isbn = {9781450367752},
publisher = {Association for Computing Machinery},
address = {New York, NY, USA},
url = {https://doi.org/10.1145/3325424.3329664},
doi = {10.1145/3325424.3329664},
booktitle = {The 5th ACM Workshop on Wearable Systems and Applications},
pages = {59–64},
numpages = {6},
location = {Seoul, Republic of Korea},
series = {WearSys '19}
}

@String{Computing = "Computing" }

@String{Computer = "{IEEE} Computer" }

@String{Academic = "Academic Press" }

@String{Springer = "Springer-Verlag" }

@inproceedings{koike2024sprout,
  title={Sprout: Designing Expressivity for Robots Using Fiber-Embedded Actuator},
  author={Koike, Amy and Wehner, Michael and Mutlu, Bilge},
  booktitle={Proceedings of the 2024 ACM/IEEE International Conference on Human-Robot Interaction},
  pages={403--412},
  year={2024}
}

@inproceedings{koike2024sprout2,
  title={Sprout: Demonstration of Soft Expressive Robot},
  author={Koike, Amy and Wehner, Michael and Mutlu, Bilge},
  booktitle={Companion of the 2024 ACM/IEEE International Conference on Human-Robot Interaction},
  pages={83--84},
  year={2024}
}

@article{rossiter2016soft,
  title={Soft robotics-The next industrial revolution?},
  author={Rossiter, Jonathan M and Hauser, Helmut},
  journal={IEEE Robotics and Automation Magazine},
  volume={23},
  number={3},
  pages={17--20},
  year={2016},
  publisher={Institute of Electrical and Electronics Engineers (IEEE)}
}

@article{laschi2016soft,
  title={Soft robotics: Technologies and systems pushing the boundaries of robot abilities},
  author={Laschi, Cecilia and Mazzolai, Barbara and Cianchetti, Matteo},
  journal={Science robotics},
  volume={1},
  number={1},
  pages={eaah3690},
  year={2016},
  publisher={American Association for the Advancement of Science}
}

@article{pfeifer2012challenges,
  title={The challenges ahead for bio-inspired'soft'robotics},
  author={Pfeifer, Rolf and Lungarella, Max and Iida, Fumiya},
  journal={Communications of the ACM},
  volume={55},
  number={11},
  pages={76--87},
  year={2012},
  publisher={ACM New York, NY, USA}
}

@article{arnold2017tactile,
  title={The tactile ethics of soft robotics: Designing wisely for human--robot interaction},
  author={Arnold, Thomas and Scheutz, Matthias},
  journal={Soft robotics},
  volume={4},
  number={2},
  pages={81--87},
  year={2017},
  publisher={Mary Ann Liebert, Inc. 140 Huguenot Street, 3rd Floor New Rochelle, NY 10801 USA}
}

@article{sun2022softsar,
  title={Softsar: The new softer side of socially assistive robots—soft robotics with social human--robot interaction skills},
  author={Sun, Yu-Chen and Effati, Meysam and Naguib, Hani E and Nejat, Goldie},
  journal={Sensors},
  volume={23},
  number={1},
  pages={432},
  year={2022},
  publisher={MDPI}
}

@article{christiansen2023brings,
  title={“It Brings the Good Vibes”: Exploring Biomorphic Aesthetics in the Design of Soft Personal Robots},
  author={Christiansen, Mads Bering and Rafsanjani, Ahmad and J{\o}rgensen, Jonas},
  journal={International Journal of Social Robotics},
  pages={1--21},
  year={2023},
  publisher={Springer}
}

@article{jorgensen2022soft,
  title={Is a soft robot more “natural”? Exploring the perception of soft robotics in human--robot interaction},
  author={J{\o}rgensen, Jonas and Bojesen, Kirsten Borup and Jochum, Elizabeth},
  journal={International Journal of Social Robotics},
  volume={14},
  number={1},
  pages={95--113},
  year={2022},
  publisher={Springer}
}

@article{christiano2017deep,
  title={Deep reinforcement learning from human preferences},
  author={Christiano, Paul F and Leike, Jan and Brown, Tom and Martic, Miljan and Legg, Shane and Amodei, Dario},
  journal={Advances in neural information processing systems},
  volume={30},
  year={2017}
}

@article{wagner2018making,
  title={Making social objects: The theory of social representation},
  author={Wagner, Wolfgang and Kello, Katrin and R{\"a}mmer, Andu},
  journal={The Cambridge handbook of sociocultural psychology},
  pages={130--147},
  year={2018},
  publisher={Cambridge University Press Cambridge, United Kingdom}
}

@inproceedings{farhadi2022exploring,
  title={Exploring the interaction kinesics of a soft social robot},
  author={Farhadi, Ulrich and Klausen, Troels Aske and J{\o}rgensen, Jonas and Vlachos, Evgenios},
  booktitle={International Conference on Human-Computer Interaction},
  pages={292--299},
  year={2022},
  organization={Springer}
}

@article{karunaratne2018exploring,
  title={Exploring fashion semiotics: communicating meanings in fashion in dress literature review},
  author={Karunaratne, Priyanka Virajini Medagedara},
  year={2018},
  publisher={South Eastern University of Sri Lanka, University Park, Oluvil, Sri Lanka.}
}

@article{klussman2020connection,
  title={Connection and disconnection as predictors of mental health and wellbeing},
  author={Klussman, Kristine and Nichols, Austin Lee and Langer, Julia and Curtin, Nicola},
  journal={International Journal of Wellbeing},
  volume={10},
  number={2},
  year={2020}
}

@article{elango2015review,
  title={A review article: investigations on soft materials for soft robot manipulations},
  author={Elango, N and Faudzi, Ahmad Athif Mohd},
  journal={The International Journal of Advanced Manufacturing Technology},
  volume={80},
  pages={1027--1037},
  year={2015},
  publisher={Springer}
}

@article{zhu2022soft,
  title={Soft, wearable robotics and haptics: Technologies, trends, and emerging applications},
  author={Zhu, Mengjia and Biswas, Shantonu and Dinulescu, Stejara Iulia and Kastor, Nikolas and Hawkes, Elliot Wright and Visell, Yon},
  journal={Proceedings of the IEEE},
  volume={110},
  number={2},
  pages={246--272},
  year={2022},
  publisher={IEEE}
}

@article{yumbla2021human,
  title={Human assistance and augmentation with wearable soft robotics: a literature review and perspectives},
  author={Yumbla, Emiliano Quinones and Qiao, Zhi and Tao, Weijia and Zhang, Wenlong},
  journal={Current Robotics Reports},
  pages={1--15},
  year={2021},
  publisher={Springer}
}

@inproceedings{o2017soft,
  title={A soft wearable robot for the shoulder: Design, characterization, and preliminary testing},
  author={O'Neill, Ciar{\'a}n T and Phipps, Nathan S and Cappello, Leonardo and Paganoni, Sabrina and Walsh, Conor J},
  booktitle={2017 International Conference on Rehabilitation Robotics (ICORR)},
  pages={1672--1678},
  year={2017},
  organization={IEEE}
}

@inproceedings{porter2020soft,
  title={Soft exoskeleton knee prototype for advanced space suits and planetary exploration},
  author={Porter, Allison P and Marchesini, Barnaba and Potryasilova, Irina and Rossetto, Enrico and Newman, Dava J},
  booktitle={2020 IEEE Aerospace Conference},
  pages={1--13},
  year={2020},
  organization={IEEE}
}

@inproceedings{pulvirenti2022towards,
  title={Towards a soft exosuit for hypogravity adaptation: Design and control of lightweight bubble artificial muscles},
  author={Pulvirenti, Emanuele and Diteesawat, Richard S and Hauser, Helmut and Rossiter, Jonathan},
  booktitle={2022 IEEE 5th International Conference on Soft Robotics (RoboSoft)},
  pages={651--656},
  year={2022},
  organization={IEEE}
}





\end{document}